\pdfoutput=1

\documentclass[10pt,twocolumn,letterpaper]{article}

\usepackage[pagenumbers]{cvpr}

\makeatletter\renewcommand\paragraph{\@startsection{paragraph}{4}{\z@}{.5em}{-1em}{\normalfont\normalsize\bfseries}}\makeatother

\usepackage{xspace}
\newcommand{\method}{MASAR\xspace}

\usepackage{booktabs}
\usepackage{pifont}

\newcommand{\std}[1]{{\tiny$\pm$#1}}
\usepackage{colortbl}
\usepackage{tikz}
\usetikzlibrary{arrows.meta}
\usepackage{float}
\ifdefined\MASARmainonly\else\usepackage{stfloats}\fi
\floatstyle{ruled}
\newfloat{algorithm}{t}{loa}
\floatname{algorithm}{Algorithm}
\floatstyle{plain}
\definecolor{oursblue}{RGB}{226,238,252}
\newcommand{\oursrow}{\rowcolor{oursblue}}
\definecolor{refgray}{gray}{0.92}
\newcommand{\refrow}{\rowcolor{refgray}}
\definecolor{linkblue}{rgb}{0.21,0.49,0.74}
\usepackage[pagebackref,breaklinks,colorlinks,allcolors=linkblue]{hyperref}
\hypersetup{pdftitle={Let Expert Confidence Guide LoRA Merging: Margin-Aware Spectral Adaptation}, pdfauthor={Alexandre Molohio, Gabriel Azpitarte, Zakaria El Ouahidi, Weiyue Li, Yan Luo, Mengyu Wang}}

\title{Let Expert Confidence Guide LoRA Merging: \\
Margin-Aware Spectral Adaptation}

\author{%
Alexandre Molohio$^{1,2}$\quad
Gabriel Azpitarte$^{1,3}$\quad
Zakaria El Ouahidi$^{1,4}$\\
Weiyue Li$^{1}$\quad
Yan Luo$^{1}$\quad
Mengyu Wang$^{1}$\\[0.5em]
$^{1}$Harvard University\quad
$^{2}$Sorbonne Universit\'e\quad
$^{3}$\'Ecole Polytechnique\quad
$^{4}$CentraleSup\'elec
}

\begin{document}
\maketitle
\begin{abstract}
Model merging combines LoRA experts specialized in different tasks into a single model without retraining, at the cost of interference between them. Recent methods reduce this interference by flattening the spectrum of each LoRA's weights, which can also weaken what each expert has learned. We show that flattening should adapt to the confidence of the experts. Less confident experts merge best when fully flattened, while more confident experts benefit from keeping more of their spectrum. We introduce MASAR, which measures this confidence with the margin, defined as the difference between an expert's top two class scores. Our method needs only one unlabeled example per task to compute this margin and set the amount of flattening. Since one example covers few input directions, we complement it with those encoded in the LoRA weights. This yields a closed-form merge that needs no labels, no training and no hyperparameter search. MASAR thus adapts to experts trained with different levels of confidence, where uniform flattening does not. On CLIP ViT-B/32, MASAR outperforms the best label-free baseline by more than 7 accuracy points. It even surpasses merges tuned on labeled data and gives the best label-free average on both ViT-L/14 and language tasks with Llama-3-8B. Our code is available at \href{https://github.com/Harvard-AI-and-Robotics-Lab/masar-merging}{\nolinkurl{github.com/Harvard-AI-and-Robotics-Lab/masar-merging}}.
\end{abstract}

\section{Introduction}
\label{sec:intro}

As pretrained models grow, fine-tuning all their weights for each new task becomes costly in both compute and storage.
Low-rank adaptation (LoRA)~\cite{lora} cuts this cost by freezing these weights and training only small low-rank
updates added to them, which makes task-specific experts light enough to share widely. Merging them gives one model
with all their capabilities, without joint retraining~\cite{modelsoups,fisher,lorahub}. The difficulty is
interference. Once the updates are added to the same weights, each expert can change the others' predictions.

Recent merges reduce this interference by flattening the singular values of the updates~\cite{isoc,dc_merge}. To understand why, we examine the eight public LoRA experts released with KnOTS~\cite{knots} for CLIP ViT-B/32. Most interference comes from the component associated with the largest singular value in each update's decomposition, which we call its top direction (\cref{fig:direction1}). But this direction also carries learned information, so attenuating it weakens the expert. This creates a trade-off; existing merges either fix the amount of flattening across banks or tune it on labeled data.

We find that the right amount of flattening for a bank depends on how confident its experts are. To vary this confidence, we train LoRA experts on the same eight tasks with cross-entropy at several logit scales~\cite{clip} and with CosFace~\cite{cosface}. We measure each expert's confidence by its prediction margin, defined as the gap between its two highest cosine scores. The KnOTS experts have the smallest margins and benefit most from full flattening, as their predictions are easily changed by other experts. Experts with larger margins are more resistant to interference, so for them full flattening removes more skill than it protects (\cref{fig:banks}b). The training recipe also changes the ranking of existing merges (\cref{fig:banks}a). How much to flatten should thus follow the margin, and measuring it takes only one unlabeled example per task.

The rest of the merge should then rely on this same example, without labels. Existing merges do not allow this.
Weight-only methods, such as Task Arithmetic (TA), TIES, TSV-M and KnOTS, compute the merge from the experts
alone~\cite{taskarithmetic,ties,tsv,knots}. Yet the weights do not show how each expert will behave once merged, and
even orthogonal updates can interfere~\cite{knots,osrm}. These methods therefore tune a coefficient that scales the
experts by grid search on labeled examples~\cite{taskarithmetic,ties,isoc}. AdaMerging and TARA do without labels, but
learn such coefficients by gradient descent, minimizing the entropy of the merged predictions~\cite{adamerging,tara}.
RegMean~\cite{regmean} needs neither labels nor gradient steps. It merges the experts in closed form, so that each
merged layer reproduces the outputs of every expert on a few unlabeled examples of its task. However, a single example per task covers only a small fraction of the input directions of each module, so even the improved variant RegMean++~\cite{regmeanpp} falls far below the individual experts (\cref{tab:vitb32}).

The experts' weights can widen the input coverage, \ie, the set of input directions on which the merge is
constrained. The right
singular vectors of each LoRA update are the directions on which that expert acts, so adding them to the
statistics of the example constrains the merge on the directions that it misses (\cref{tab:ingredients}). ACTMat, WUDI, ACE-Merging and ReForge also do so~\cite{actmat,wudi,acemerging,reforge}, but from the raw updates, whose top direction dominates what they infer. We read them from the flattened updates instead,
so the margin that sets the flattening also shapes this completion.

Building on these two observations, we introduce \method (Margin-Aware Spectral Adaptation and Reconstruction). It
measures the margin of each expert relative to that of the pretrained model on the same example, and keeps more of
each spectrum as this relative margin grows. Flattening at fixed norm also changes how strongly each
expert responds to its inputs. \method then measures this change and rescales each expert to restore its response. Finally, it merges each module in closed form, as RegMean does, but with the example
completed by the input directions of the weights (\cref{fig:framework}). All its hyperparameters thus come from one
unlabeled example per task, with no gradient step or search, and the same rules apply to vision and language models.
Our contributions are:
\begin{itemize}
    \item We show that how much to flatten LoRA experts before merging depends on how they were trained, and that
    their prediction margin on one image indicates it. Setting the spectrum from this margin outperforms both full
    flattening and no flattening on average over 14 banks (\cref{sec:margins}).
    \item We show that the LoRA weights complete the activation statistics of one example, best through the flattened
    targets rather than the raw updates (\cref{sec:reconstruction}).
    \item We introduce \method and evaluate it on CLIP ViT-B/32 and ViT-L/14 and on six NLI experts of Llama-3-8B.
    With one example per task, it gives the best label-free average on all three, and outperforms or matches the merges
    that tune their scale on labels (\cref{sec:experiments}). RegMean++~\cite{regmeanpp} needs several images to beat
    it.
\end{itemize}

\section{Related work}
\label{sec:related}

\paragraph{Merging with activations.}
Some merges learn their coefficients by gradient descent on unlabeled inputs, minimizing the prediction entropy for AdaMerging and
TARA~\cite{adamerging,tara} and the null-space ratio of the LoRA inputs for NSC~\cite{nsc}. Others reconstruct
the experts' outputs, as RegMean does in closed form~\cite{regmean}. RegMean++ and Chain of Merges collect its inputs through the partially merged
model~\cite{regmeanpp,chainofmerges}, CrossMean instead pairs merged and expert activations~\cite{crossmean}, and
IterIS alternates between inference and solving the regression to merge LoRAs from few samples~\cite{iteris}. Without
inputs, ACTMat, WUDI and ACE-Merging estimate each expert's input statistics from the updates
themselves~\cite{actmat,wudi,acemerging}. Finally, ReForge mixes activation and weight Grams with one or five examples
per task, and tunes the mix on labels~\cite{reforge}. \method also completes the example with the weights, but from
the flattened targets and without labels.

\paragraph{Spectral merging.}
Some spectral merges choose the directions along which to merge. TSV-M keeps the top singular vectors of each task, orthogonalized across tasks~\cite{tsv}, KnOTS merges LoRA
updates in the SVD basis of their concatenation, where they align better~\cite{knots}, and Core Space in a low-rank
basis shared by all tasks, without loss of information~\cite{corespace}. Iso-CTS~\cite{isoc} and
CT-Merging~\cite{ctmerging} instead separate shared and task-specific directions, which CT-Merging scales per
task. Others reshape the singular values. Iso-C sets those of the summed update to their
mean~\cite{isoc}, ACE-Merging smooths its merged spectrum only when the norms of the updates differ
strongly~\cite{acemerging}, and DC-Merge, closest to our spectral step, flattens each expert's spectrum, as a few
singular values dominate each update~\cite{dc_merge}. None of them reads the experts' confidence. \method instead flattens
according to their prediction margins, fully when margins are small and less as they grow.

\paragraph{Expert confidence and margins.}
How experts were trained also affects merging~\cite{tangent,overtraining}. DisTaC finds that low-confidence experts,
for instance trained with label smoothing, degrade merges. It thus distills them on unlabeled data to raise their
confidence~\cite{distac}. MaLoRA, from a linear margin analysis, weights each adapter inversely to its margin on
labeled data~\cite{malora}. FeatCal also relates the score drift of a merged model to the expert's
margin~\cite{featcal}. In contrast, \method neither retrains the experts nor uses labels, and reads their margins on one unlabeled
example, relative to those of the pretrained model. These margins are all it needs to adapt the flattening to how the experts were trained.

\section{Spectral reconstruction from one image}
\label{sec:reconstruction}

\providecommand{\seclink}[5]{%
  \node[anchor=north west, inner sep=0] at (#1,#2)
    {\hyperref[#5]{\phantom{\rule{#3\textwidth}{#4\textwidth}}}};}

\begin{figure*}[t]
  \centering
  \begin{tikzpicture}
    \node[anchor=south west, inner sep=0] (img) at (0,0)
      {\includegraphics[width=0.895\textwidth]{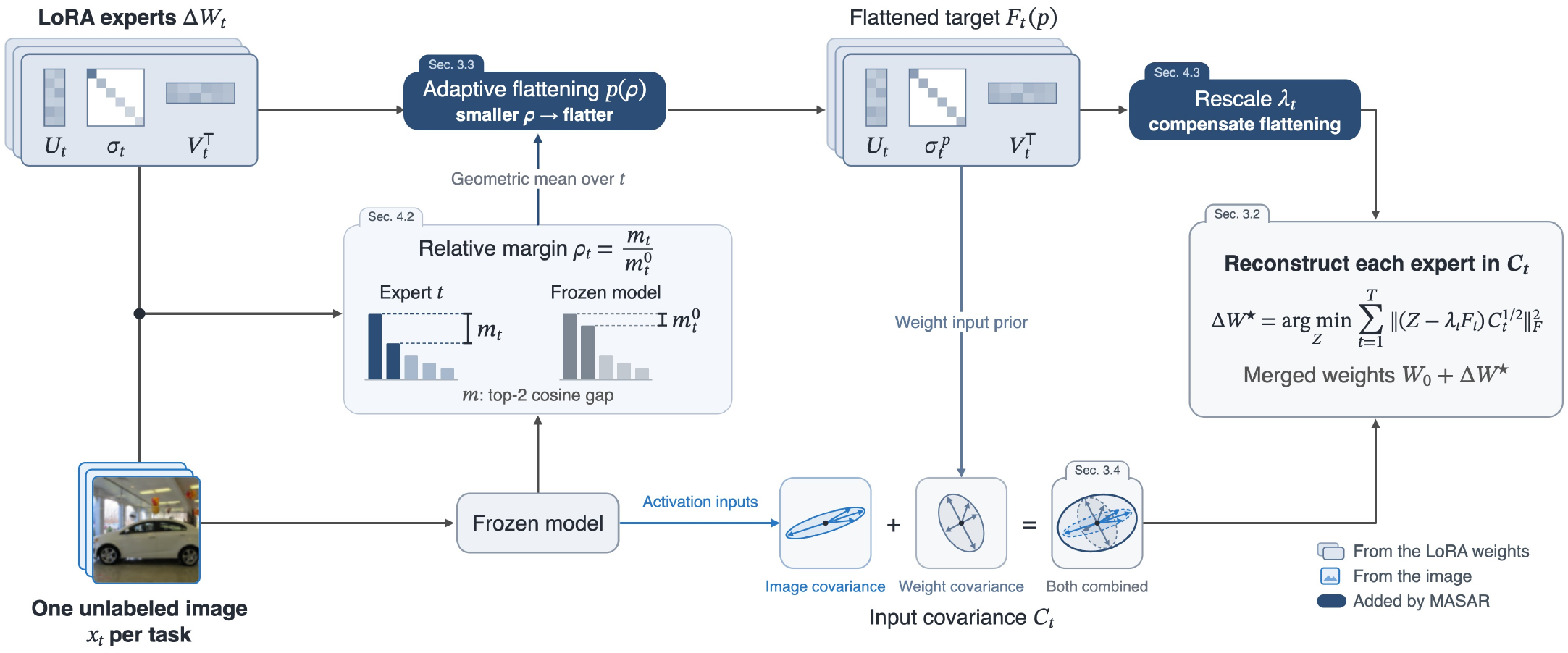}};
    \begin{scope}[x={(img.south east)}, y={(img.north west)}]
      \seclink{0.257}{0.917}{0.168}{0.048}{sec:flatten}  %
      \seclink{0.720}{0.905}{0.133}{0.050}{sec:rescale}    %
      \seclink{0.219}{0.683}{0.248}{0.132}{sec:p-rule}     %
      \seclink{0.758}{0.688}{0.239}{0.136}{sec:objective}  %
      \seclink{0.497}{0.295}{0.231}{0.069}{sec:metric}     %
    \end{scope}
  \end{tikzpicture}
  \caption{\textbf{\method overview.} With one unlabeled image per task, the geometric
mean of expert relative margins $\rho_t=m_t/m_t^0$ sets $p$. Each update
$\Delta W_t$ is flattened at fixed norm into $F_t(p)$, then rescaled by
$\lambda_t$. The metric $C_t$ combines image-token and weight-input-direction
covariances. The merged update $Z$ must match each $\lambda_t F_t$ primarily
along these directions, yielding $\Delta W^\star$. Section labels are clickable.}
  \label{fig:framework}
\end{figure*}

We first describe the setting (\cref{sec:setting}) and write the merge as one reconstruction (\cref{sec:objective}). We then flatten each expert (\cref{sec:flatten}) and finally complete the image with the weights (\cref{sec:metric}). \Cref{fig:framework} gives an overview of \method.

\subsection{Setting}
\label{sec:setting}

\paragraph{Experts.} We merge a bank of $T$ LoRA experts, one per task. All are fine-tuned from the same pretrained model. In every attention block, LoRA adapts four weight matrices: the query, key, value and output projections. We call each of them a module, and index modules by $\ell\in\{1,\dots,L\}$. The merge is the same for every module, so we describe it for one and omit the index $\ell$. This module has pretrained weight $W_0\in\mathbb{R}^{d_{\mathrm{out}}\times d}$. Expert $t\in\{1,\dots,T\}$ adds to it the LoRA update $\Delta W_t=B_tA_t$ of rank
$r=16$, with $B_t\in\mathbb{R}^{d_{\mathrm{out}}\times r}$ and $A_t\in\mathbb{R}^{r\times d}$. Its singular value decomposition is
\begin{equation}
    \Delta W_t=U_t\operatorname{diag}(\sigma_t)V_t^\top=\textstyle\sum_{j=1}^{r}\sigma_{t,j}\,u_{t,j}v_{t,j}^\top,
    \label{eq:svd}
\end{equation}
    with singular values $\sigma_{t,1}\geq\dots\geq\sigma_{t,r}$ and orthonormal singular vectors $u_{t,j}$ and $v_{t,j}$. The update is thus a sum of $r$ rank-one matrices. Moreover, $\Delta W_tx$ depends on an input $x$ only through its projections on the $v_{t,j}$, which we call its read directions.

\paragraph{Access.} \method receives the bank of experts, the frozen pretrained model, the classification head of each task and one unlabeled training image $x_t$ per task. It uses no labels, no gradient steps and no hyperparameter search, and returns one model without a router. When the pretrained model processes $x_t$, the module receives $n$ tokens, the columns of $X_t\in\mathbb{R}^{d\times n}$. Their uncentered second moment $\hat\Sigma_t=\frac1nX_tX_t^\top$ gives their mean squared projection on each input direction. On ViT-B/32, $n=50$ and $d=768$ ($n=257$ and $d=1024$ on ViT-L/14). The image thus constrains the merge on at most 50 of the 768 input directions.

\subsection{Merging by reconstruction}
\label{sec:objective}

RegMean~\cite{regmean} merges the experts in closed form. It looks for the merged weight $W_0+Z$ that best reproduces the outputs of every expert on its own inputs $\tilde X_t$:
\begin{equation}
    \begin{gathered}
        \min_Z\sum_{t=1}^{T}\big\|(Z-\Delta W_t)\,\tilde G_t^{1/2}\big\|_F^2,\\
        \Delta W^{\mathrm{RM}}=\Big(\sum_{t=1}^{T}\Delta W_t\tilde G_t\Big)\Big(\sum_{t=1}^{T}\tilde G_t\Big)^{-1},
    \end{gathered}
    \label{eq:regmean}
\end{equation}
where $\tilde G_t$ is the Gram matrix $\tilde X_t\tilde X_t^\top$ of these inputs, with off-diagonal entries scaled by $\alpha=0.9$. With one image, however, these inputs span at most $n$ input directions. Outside them, the merge is constrained only by the diagonal of $\tilde G_t$, which depends on the image alone, not on the expert. 

We keep this reconstruction but make three choices:
\begin{equation}
    \begin{gathered}
        \min_Z\sum_{t=1}^{T}\big\|(Z-\lambda_tF_t)\,C_t^{1/2}\big\|_F^2,\\
        \Delta W^\star=\Big(\sum_{t=1}^{T}\lambda_tF_tC_t\Big)\Big(\sum_{t=1}^{T}C_t\Big)^{-1}.
    \end{gathered}
    \label{eq:objective}
\end{equation}
First, the merge reproduces a flattened target $F_t$ instead of each update (\cref{sec:flatten}). Second, a metric $C_t$ replaces $\tilde G_t$ to decide which input directions the merge must reproduce well. Unlike $\tilde G_t$, it completes the image with the weights (\cref{sec:metric}). Finally, $\lambda_t>0$ scales expert $t$. Both $F_t$ and $C_t$ depend on the amount of flattening (Supp.~\ref{sec:supp-decoupling}). It is set by an exponent $p$ (\cref{sec:flatten}), chosen from the experts' margins in \cref{sec:p-rule}, and omitted when fixed. The solution $\Delta W^\star$ is cheap: one linear solve per module, by Cholesky factorization (Supp.~\ref{sec:supp-cost}). It stays low-rank, with rank at most $Tr$, and is added directly to $W_0$ (Supp.~\ref{sec:supp-math-reconstruction}). To isolate these choices from the scale, this section uses one $\lambda$ chosen on labels; \cref{sec:rescale} then sets each $\lambda_t$ without labels.

\subsection{Flattening each expert}
\label{sec:flatten}

\begin{figure}[H]
    \centering
    \includegraphics[width=\linewidth]{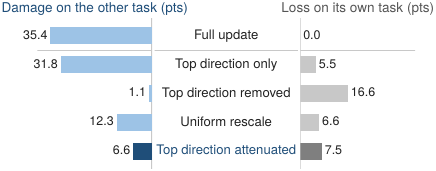}
    \caption{\textbf{The top direction carries both the interference and the skill.} KnOTS ViT-B/32, validation accuracy. The update of one
    expert is added to another expert (left: accuracy lost on that expert's task) or used alone
    (right: accuracy lost on its own task); mean over the 56 ordered pairs of the eight experts. \emph{Uniform rescale}: the whole update scaled to the norm of the attenuated one. \emph{Attenuated}: $\sigma_{t,1}$
    lowered to the root mean square of $\sigma_t$.}
    \label{fig:direction1}
\end{figure}

On KnOTS, the top direction of each expert carries almost all the interference (\cref{fig:direction1}).
Attenuating it, rather than deleting it or scaling down the whole update, keeps both this interference and the
expert's own loss small. We therefore flatten each spectrum, which shrinks its top directions relative to the weak
ones, and keep its norm. For $p\in[0,1]$, the target is
\begin{equation}
    \begin{gathered}
        F_t(p)=c_t(p)\,U_t\operatorname{diag}\big(\sigma_t^{\,p}\big)V_t^\top,\\
        c_t(p)=\frac{\|\sigma_t\|_2}{\|\sigma_t^{\,p}\|_2}.
    \end{gathered}
    \label{eq:target}
\end{equation}
Each singular value is raised to the power $p$, and $c_t(p)$ restores the norm of $\Delta W_t$. The singular vectors do not change. Thus $p=1$ gives $\Delta W_t$, and $p=0$ makes all singular values equal, which gives a flat spectrum. A smaller $p$ flattens more. Since the norm is kept, $p$ sets the shape of the spectrum and $\lambda_t$ its scale.

Full flattening ($p=0$) gives the flat-spectrum update closest to the expert in Frobenius norm (Supp.~\ref{sec:supp-math-flat}). It removes the dominance of the top direction without dropping any direction. On KnOTS, flatter targets merge better, from 61.3\% at $p=1$ to 67.8\% at $p=0$.

\subsection{Completing the image with the weights}
\label{sec:metric}

The image covers only a few input directions, and these are not always the read directions of the expert. On KnOTS, the image covers the top one, $v_{t,1}$, with 7.1 times the mean squared projection of an average input direction. However, it misses the other fifteen. Only 7\% of their squared norm lies in the span of its 50 tokens (median), about as much as for a random direction (6.5\%). We therefore complete the image with a matrix built from the weights, the weight input prior
\begin{equation}
    \begin{gathered}
        P_t(p)=V_t\operatorname{diag}\big(w_t(p)\big)V_t^\top+\varepsilon I,\\
        w_{t,j}(p)=\frac{r\,\sigma_{t,j}^{2p}}{\sum_{i=1}^{r}\sigma_{t,i}^{2p}},
    \end{gathered}
    \label{eq:prior}
\end{equation}
with $\varepsilon=0.01$. The first term of $P_t$ is the normalized Gram matrix of the flattened target $F_t$ (Supp.~\ref{sec:supp-math-reconstruction}). At $p=0$, it gives all read directions the same weight. ACTMat and ReForge use the Gram matrix of the raw update instead~\cite{actmat,reforge}, which the top direction dominates. The metric $C_t$ then averages the image's second moment $\hat\Sigma_t$ and the prior $P_t$, both scaled to trace $d$ (other mixes in Supp.~\ref{sec:supp-sensitivity}):
\begin{equation}
    \begin{gathered}
        C_t(p)=\tfrac12\,\mathcal{N}\big(\hat\Sigma_t\big)+\tfrac12\,\mathcal{N}\big(P_t(p)\big),\\
        \mathcal{N}(A)=\frac{d\,A}{\operatorname{tr}A}.
    \end{gathered}
    \label{eq:metric}
\end{equation}
It penalizes the error on the image's tokens and the read directions. The term $\varepsilon I$ adds a small weight to every input direction, which makes $C_t$ positive definite. At $p=0$, the fifteen directions that the image missed now get 17 times the weight of an average input direction (median), against 0.9 for $\mathcal{N}(\hat\Sigma_t)$ alone. We measure $\hat\Sigma_t$ in the frozen pretrained model, not in the expert (RegMean) or the partially merged model (RegMean++~\cite{regmeanpp}), so only once per image.

\begin{table}[t]
  \centering
  \footnotesize
  \setlength{\tabcolsep}{4pt}
  \begin{tabular}{@{}l c c@{}}
    \toprule
    Variant & Acc. & $\Delta$ \\
    \midrule
    \oursrow \textbf{Full reconstruction} & \textbf{67.8}\std{0.8} & -- \\
    \midrule
    $-$ flattening ($p=1$) & 61.3\std{0.5} & $-6.6$ \\
    $-$ weight input prior (image moment only) & 62.4\std{1.3} & $-5.4$ \\
    \quad weight input prior $\to$ identity & 61.4\std{1.1} & $-6.4$ \\
    \quad weight input prior $\to$ raw update~\cite{actmat,reforge} & 67.0\std{1.1} & $-0.8$ \\
    $-$ image (weight input prior only) & 63.8 & $-4.0$ \\
    \midrule
    RegMean~\cite{regmean} (raw targets, expert inputs) & 56.4\std{0.4} & $-11.5$ \\
    \bottomrule
  \end{tabular}
  \caption{\textbf{The weight input prior completes one image.} Ingredients of the reconstruction on KnOTS ViT-B/32, with one unlabeled image per task and $p=0$. Each row removes one ingredient; one $\lambda$ shared by all experts is tuned on labeled images, as for the baselines. Test accuracy (\%), mean $\pm$ std over five image draws.}
  \label{tab:ingredients}
\end{table}

The image and the prior are both needed (\cref{tab:ingredients}). On KnOTS, the image alone gives 62.4\% and the prior alone 63.8\%, against 67.8\% together. An identity matrix in place of the prior gives only 61.4\%, so the gain does come from the weights. Finally, building the prior from the raw update, as ACTMat and ReForge do, gives 67.0\%. Building it from the flattened target thus adds 0.8 points, on each of the five image draws.

\section{Choosing $p$ and $\lambda_t$ without labels}
\label{sec:margins}

We relate gains from flattening to expert margins
(\cref{sec:banks}), then use the same unlabeled image per task
to choose $p$ from these margins (\cref{sec:p-rule})
and adjust each expert's scale $\lambda_t$ (\cref{sec:rescale}),
as shown in \cref{fig:framework}.
\Cref{alg:masar} summarizes \method.

\subsection{Experts trained differently merge differently}
\label{sec:banks}

KnOTS is a single bank, so we train new banks ourselves, from the same pretrained model and on the same eight tasks,
with one recipe in which only the loss changes. Let $z_c$
be the cosine score of class $c$ in the task head, $y$ the true class and $s$ a scale. The cross-entropy (CE) and
CosFace (CF)~\cite{cosface} losses are
\begin{equation}
    \begin{gathered}
        \mathcal{L}_{\mathrm{CE}}=-\log\frac{e^{s z_y}}{\sum_{c}e^{s z_c}},\\
        \mathcal{L}_{\mathrm{CF}}=-\log\frac{e^{s(z_y-\mu)}}{e^{s(z_y-\mu)}+\sum_{c\neq y}e^{s z_c}}.
    \end{gathered}
    \label{eq:losses}
\end{equation}
For CE, decreasing $s$ encourages larger cosine-score gaps
to achieve the same loss. CF instead subtracts a margin
$\mu$ from the true-class score, encouraging gaps beyond $\mu$.
We vary $s$ for CE and $\mu$ for CF at fixed $s=10\,000$
to obtain experts with different prediction margins.

\begin{figure}[t]
    \centering
    \includegraphics[width=\linewidth]{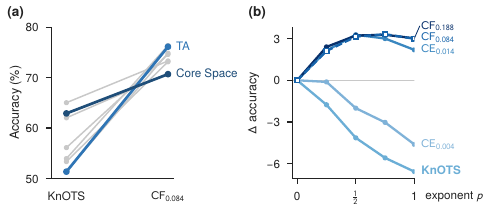}
    \caption{\textbf{Training changes merge rankings and the preferred
degree of flattening.}
(a) Test accuracy of the eight weight-only merges in Tab.~2.
Each line connects the same method on KnOTS and
$\mathrm{CF}_{0.084}$.
(b) Change in test accuracy relative to full flattening
($p=0$), with $\lambda$ tuned on labeled images.
Positive values indicate a gain from partial or no flattening.}
    \label{fig:banks}
\end{figure}

We measure expert $t$'s confidence by its margin $m_t$, the gap
between its two highest cosine scores on one unlabeled image $x_t$.
The margin $m$ of a bank is the median of the $m_t$, and we denote banks by
$\mathrm{CE}_m$ or $\mathrm{CF}_m$, such as $\mathrm{CF}_{0.084}$. We train eight banks on ViT-B/32 and four on
ViT-L/14; with the public KnOTS bank of each backbone, this gives the 14 banks listed in \cref{sec:setup}.

The KnOTS experts are barely confident, with the smallest margins of all banks ($m=0.002$ on ViT-B/32 and 0.004
on ViT-L/14). We observe that this changes how they merge. Every weight-only merge does better on the confident
experts of $\mathrm{CF}_{0.084}$, but unequally (\cref{fig:banks}a). Core Space drops from second to last and TA rises
from last to second, so a ranking obtained on KnOTS alone does not carry over to other experts. The best spectrum also follows the margin
(\cref{fig:banks}b). Small-margin banks such as KnOTS are best fully flattened ($p=0$), whereas
large-margin banks gain about 3 points by keeping part of their spectrum ($p$ between $\frac12$ and $\frac34$). This also occurs among experts trained with cross-entropy. $\mathrm{CE}_{0.004}$ and $\mathrm{CE}_{0.014}$ are trained with the
same cross-entropy, but the smaller scale $s$ of $\mathrm{CE}_{0.014}$ (1\,000 instead of 10\,000) gives it larger
margins, and only $\mathrm{CE}_{0.014}$ benefits from keeping part of the spectrum.

We test how interference affects banks with different margins.
In Task Arithmetic, we add only the top directions of the other
experts to each expert. On validation images, accuracy drops
by 14.2 points on KnOTS and 19.4 on $\mathrm{CE}_{0.004}$,
but does not drop on the large-margin banks $\mathrm{CE}_{0.084}$
and $\mathrm{CF}_{0.084}$. Flattening weakens these top
directions (\cref{sec:flatten}).

\subsection{Choosing $p$ from the relative margin}
\label{sec:p-rule}

\begin{figure*}[t]
    \centering
    \includegraphics[width=\textwidth]{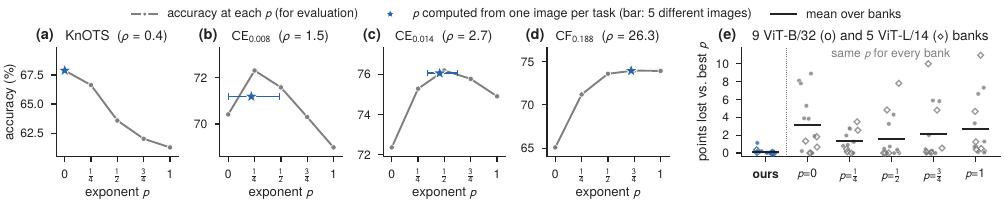}
    \caption{\textbf{Choosing the flattening exponent from relative margins.}
(a--d) Test accuracy on four ViT-B/32 banks ordered by $\rho$.
Gray curves evaluate fixed $p$ values; blue stars mark the
choice from \cref{eq:p-rule}. $p=0$ gives full flattening
and $p=1$ no flattening. Horizontal bars show variation in
the selected $p$ across five draws of one image per task.
(e) Accuracy points lost relative to each bank's best $p$
on the tested grid. Each marker represents a bank
(circles: ViT-B/32; diamonds: ViT-L/14); black bars show means.
``Ours'' selects $p$ per bank; each other column uses one
fixed $p$ for all banks. Lower is better.}
    \label{fig:margin-rule}
\end{figure*}

The results in \cref{fig:banks}b motivate choosing $p$ from expert margins. However, the same training recipe gives margins up to twice as large on ViT-L/14 as on ViT-B/32. To account for this difference, we compare each expert's margin with that of its pretrained model on the same image. Let $m^0_t$ be the gap between the pretrained model's two highest cosine scores on $x_t$, using the head of task $t$. The relative margin of expert $t$, defined when $m^0_t>0$, is
\begin{equation}
    \rho_t=\frac{m_t}{m^0_t}.
    \label{eq:rho}
\end{equation}
We take the geometric mean $\rho=(\prod_{t=1}^{T}\rho_t)^{1/T}$ and set
\begin{equation}
    p=p_{\max}\,\max\Big(0,\;1-\frac{1}{\rho}\Big),\qquad p_{\max}=\tfrac34.
    \label{eq:p-rule}
\end{equation}
When $\rho\le1$, the experts have not increased their margins over the pretrained model on average. We then set $p=0$ and flatten fully. When $\rho>1$, their margins have increased. As $\rho$ grows, $1-1/\rho$ increases, giving a larger $p$. We thus flatten less and keep more weight on the experts'
strongest directions. Even large-margin banks benefit from partial flattening (\cref{fig:banks}b), so we set $p_{\max}=\frac34$ to retain some flattening even at large relative margins. This limits the dominance of the top direction in both the target and the weight input prior. We chose $p_{\max}$ on validation images and keep it fixed across banks and models. With this setting, KnOTS gets $p=0$ ($\rho<0.4$), while the large-margin banks get $p\approx0.6$ to 0.7.

\Cref{fig:margin-rule} evaluates our rule for choosing $p$ on the test set. As the relative margin $\rho$ grows, the best tested $p$
increases from 0 to $\frac34$, and our rule gives accuracy
close to the maximum of each curve (a--d). Across the 14 banks, our rule achieves 77.5\% mean test accuracy. Choosing the best tested $p$ for each bank raises mean accuracy by only 0.15 points compared with our rule. Even the best fixed $p$, $\frac14$, loses up to 3.5 points
against each bank's best tested accuracy, compared with
at most 1.1 for our rule. The two extreme fixed choices also give lower mean accuracy than our rule's 77.5\%: 74.6\% with full flattening ($p=0$) and 75.0\% without flattening ($p=1$) (e). These averages also hide differences between banks. Compared with our rule, full flattening loses 4.2 points on average on the ten banks where we choose partial flattening. No flattening loses 7.3 points on average on the four banks where we choose full flattening.

To see how the choice of image affects $p$, we repeat the merge with new random images. Over five repeats, $p$ varies by less than 0.15 for $\rho>5$ and by at most 0.5 on $\mathrm{CE}_{0.008}$, whose $\rho$ lies around 1 (bar in \cref{fig:margin-rule}b). Over 64 repeats, the accuracy varies by at most 0.6 points (standard deviation), even on $\mathrm{CE}_{0.014}$, where $p$ ranges from 0.07 to 0.67, whereas an image of another task costs 5.3 points (Supp.~\ref{sec:supp-calibration}).

\subsection{Rescaling each expert}
\label{sec:rescale}

Flattening with the chosen $p$ preserves the weight norms,
but can change the magnitude of each target's response
by a different amount. We therefore separate each expert's
scale $\lambda_t$ in \cref{eq:objective} into two factors,
\begin{equation}
    \lambda_t=\nu\,\kappa_t(p),\qquad \nu=\tfrac12.
    \label{eq:lambda}
\end{equation}
The factor $\kappa_t(p)$ compensates for this change for
each expert. The common factor $\nu$ controls the overall
strength of the merged update, as the shared scale does
in Task Arithmetic~\cite{taskarithmetic}.

\begin{figure}[t]
    \centering
    \includegraphics[width=\linewidth]{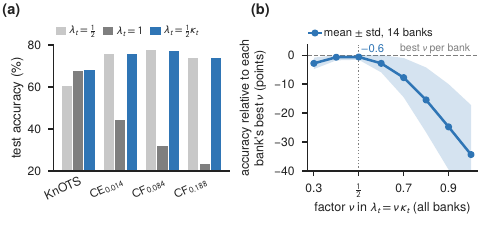}
    \caption{\textbf{With $\kappa_t$, one factor $\nu=\frac12$ suits every bank.} (a)~Test accuracy with one scale for
    every expert ($\lambda_t=\frac12$ or 1) and with $\lambda_t=\frac12\kappa_t$, on the four ViT-B/32 banks of
    \cref{tab:vitb32}. (b) Accuracy with one factor $\nu$ for all banks ($\lambda_t=\nu\kappa_t$), relative to the best tested
$\nu$ of each bank (dashed).}
    \label{fig:lambda}
\end{figure}

\paragraph{Matching response norms.} Let $X_{t\ell}$ be the tokens received by module $\ell$ when the pretrained model processes the image of task $t$. On these inputs, the expert update changes the output by $\Delta W_{t\ell}X_{t\ell}$, and the flattened target by $F_{t\ell}(p)X_{t\ell}$. The weights $\Delta W_{t\ell}$ and $F_{t\ell}(p)$ have the same norm, but flattening, which makes the target's singular values proportional to $\sigma_{t\ell,j}^{\,p}$, moves weight from the top read direction $v_{t\ell,1}$ to the weak ones. The target can thus respond more weakly than the expert on these inputs, whereas flattening should change the balance between directions, not the expert's strength. We therefore set $\kappa_{t\ell}(p)$ to the ratio of the two response norms,
\begin{equation}
    \kappa_{t\ell}(p)^2=\frac{\|\Delta W_{t\ell}X_{t\ell}\|_F^2}{\|F_{t\ell}(p)X_{t\ell}\|_F^2}
    =\frac{\sum_j\sigma_{t\ell,j}^{2}\,e_{t\ell,j}\big/\sum_j\sigma_{t\ell,j}^{2}}
    {\sum_j\sigma_{t\ell,j}^{2p}\,e_{t\ell,j}\big/\sum_j\sigma_{t\ell,j}^{2p}},
    \label{eq:kappa}
\end{equation}
where $e_{t\ell,j}=v_{t\ell,j}^\top\hat\Sigma_{t\ell}v_{t\ell,j}$ is the energy of the tokens along $v_{t\ell,j}$. On the right-hand side, both terms average the same energies
$e_{t\ell,j}$, the numerator with the weights $\sigma_{t\ell,j}^2$ and the denominator with the flatter weights
$\sigma_{t\ell,j}^{2p}$. Without flattening ($p=1$), $\kappa_{t\ell}=1$. If the tokens carry more energy along the stronger read directions, $\kappa_{t\ell}(p)\ge1$, and the flatter the target, the larger $\kappa_{t\ell}(p)$ (Supp.~\ref{sec:supp-math}).
To have one correction per expert, we take $\kappa_t(p)$
as the geometric mean of the $\kappa_{t\ell}(p)$ over
the $L$ modules. \Cref{fig:lambda}a compares $\lambda_t=\frac12\kappa_t$ with the common scales $\lambda_t=\frac12$ and $\lambda_t=1$, and neither suits every bank. The first leaves the fully flattened KnOTS experts too weak, and the second loses 31 to 51 accuracy points on the other banks against $\frac12\kappa_t$.

\begin{algorithm}[t]
    \caption{\method.}
    \label{alg:masar}
    \footnotesize
    \begin{tabular}{@{}r@{\hspace{5pt}}>{\raggedright\arraybackslash}p{0.78\linewidth}@{}r@{}}
        \multicolumn{3}{@{}p{\linewidth}@{}}{\textbf{Input:} frozen weights $W_{0\ell}$, LoRA updates $\Delta W_{t\ell}$ and one
        unlabeled image $x_t$ per task, for $T$ experts and $L$ modules.} \\
        \multicolumn{3}{@{}p{\linewidth}@{}}{\textbf{Constants:} $p_{\max}=\frac34$, $\nu=\frac12$, $\varepsilon=0.01$, the
        same for every bank.} \\
        \multicolumn{3}{@{}p{\linewidth}@{}}{\textbf{Output:} merged weights $W_{0\ell}+\Delta W^\star_\ell$.} \\[2pt]
        1: & SVD of each $\Delta W_{t\ell}$, in double precision \hfill\textit{once per bank} & (\ref{eq:svd}) \\
        2: & \textbf{for} $t=1,\dots,T$ \textbf{do} & \\
        3: & \quad Frozen model on $x_t$: tokens $X_{t\ell}$ & \\
        4: & \quad Gaps $m^0_t$, $m_t$ on $x_t$;\quad $\rho_t\leftarrow m_t/m^0_t$ & \\
        5: & $\rho\leftarrow$ geometric mean of the $\rho_t$ & (\ref{eq:rho}) \\
        6: & $p\leftarrow p_{\max}\max(0,1-1/\rho)$ & (\ref{eq:p-rule}) \\
        7: & $\lambda_t\leftarrow\nu\,\kappa_t$, with $\kappa_t=\big(\prod_{\ell}\kappa_{t\ell}(p)\big)^{1/L}$, for every $t$ & (\ref{eq:lambda})--(\ref{eq:kappa}) \\
        8: & \textbf{for} $\ell=1,\dots,L$ \textbf{do} & \\
        9: & \quad $F_{t\ell}(p)$, $P_{t\ell}(p)$ and $C_{t\ell}(p)$ for every $t$ & (\ref{eq:target})--(\ref{eq:metric}) \\
        10: & \quad $\Delta W^\star_\ell\leftarrow\big(\sum_t\lambda_tF_{t\ell}C_{t\ell}\big)\big(\sum_tC_{t\ell}\big)^{-1}$ & (\ref{eq:objective}) \\
        11: & \textbf{return} $W_{0\ell}+\Delta W^\star_\ell$ for every $\ell$ & \\
    \end{tabular}
\end{algorithm}

\paragraph{Scaling the merged update.}
Multiplying all targets by $\nu$ in \cref{eq:objective}
multiplies the merged update by the same factor.
We use $\nu=\frac12$ for every bank and model.

\Cref{fig:lambda}b evaluates this common choice on all
14 banks. No other tested value of $\nu$ comes closer
than $\nu=\frac12$ to the best factor of each bank, 0.6 points
below it on average.
The tested values below $\frac12$ change accuracy by only
a few points, while $\nu=0.7$ reduces accuracy by 7.8
points on average.

\section{Experiments}
\label{sec:experiments}

\subsection{Setup}
\label{sec:setup}

\paragraph{Experts.} We merge eight LoRA experts of rank 16, one per task (SUN397, Cars, RESISC45, EuroSAT, SVHN,
GTSRB, MNIST and DTD~\cite{sun397,cars,resisc45,eurosat,svhn,gtsrb,mnist,dtd}), on the vision encoders of CLIP ViT-B/32 and ViT-L/14~\cite{clip,vit}. Each task keeps its own text
head, as in KnOTS~\cite{knots}. We use 14 banks, named by their margin $m$ (\cref{sec:banks}): KnOTS, $\mathrm{CE}_{0.004}$, $\mathrm{CE}_{0.008}$, $\mathrm{CE}_{0.014}$, $\mathrm{CE}_{0.043}$, $\mathrm{CE}_{0.084}$, $\mathrm{CF}_{0.084}$,
$\mathrm{CF}_{0.116}$ and $\mathrm{CF}_{0.188}$ on ViT-B/32, and KnOTS, $\mathrm{CE}_{0.007}$, $\mathrm{CE}_{0.097}$, $\mathrm{CF}_{0.100}$ and $\mathrm{CF}_{0.231}$ on ViT-L/14. KnOTS, the public bank, has the smallest margins
($m=0.002$ on ViT-B/32 and 0.004 on ViT-L/14). We train the other banks as in \cref{sec:banks}, on the same modules with AdamW~\cite{adamw} (learning rate $3\cdot10^{-4}$, batch size 32, weight
decay 0.1) and early stopping (Supp.~\ref{sec:supp-banks}). Tables~\ref{tab:vitb32} and~\ref{tab:vitl14} report four and three banks with spread
margins, and \cref{fig:margin-rule}e, \cref{fig:lambda}b and \cref{tab:label-free} use all 14; Supp.~\ref{sec:supp-allbanks} gives the baselines on all 14, and Supp.~\ref{sec:supp-closest} the closest activation-based merges, tuned on labels.

\paragraph{Images and evaluation.} Each merge receives one unlabeled training image per task (Supp.~\ref{sec:supp-splits}). We report the accuracy
averaged over the eight tasks, as mean and standard deviation over five draws of this image. We developed the method
on up to 512 validation images per task, also used by \cref{fig:direction1} and the other diagnostics of
\cref{sec:reconstruction,sec:margins}. All tables and \cref{fig:banks,fig:margin-rule,fig:lambda,fig:budget}
use 1,536 images per task of the official test splits (1,024 for DTD), never used to adjust
$\lambda$.

\paragraph{Baselines.} Weight-only merges compute the merge from the LoRA weights, then scale it by one $\lambda$:
TA~\cite{taskarithmetic}, TIES~\cite{ties}, KnOTS-TIES~\cite{knots}, TSV-M~\cite{tsv}, Iso-C~\cite{isoc}, DC-Merge~\cite{dc_merge},
Core Space~\cite{corespace}, CT-Merging~\cite{ctmerging} and ACTMat~\cite{actmat}. As in KnOTS, they try on each bank every $\lambda$ of a grid (with the density for
TIES and the smoothing for DC-Merge)
and keep the most accurate on 128 labeled images per task (Supp.~\ref{sec:supp-baselines}), so they need labels (\ding{55} in the tables); for ACTMat,
this search is our addition.
Activation-based merges use no label (\checkmark) and calibrate on the same unlabeled image per task as \method.
RegMean~\cite{regmean}, RegMean++~\cite{regmeanpp} and CrossMean~\cite{crossmean} reconstruct the experts in closed
form, without a scale as in their papers, while AdaMerging~\cite{adamerging} and TARA-B~\cite{tara} learn their
coefficients with 500 unsupervised steps on the image, and NSC~\cite{nsc} with the 100 steps of its paper (500 on NLI), by
minimizing the null-space ratio of the LoRA inputs from an initial coefficient that we choose on the labeled images
(\ding{55}). \emph{Individual experts} gives the accuracy of each expert on
its own task.

\subsection{Results on ViT-B/32}
\label{sec:results-b32}

\Cref{tab:vitb32} compares all merges on the four ViT-B/32 banks. \method is the best label-free merge on every bank (all banks in Supp.~\ref{sec:supp-allbanks}),
and it also outperforms every weight-only merge, although these choose their $\lambda$ on labeled images. It reaches 73.8\% on average, against 66.4\% for
TARA-B, the best label-free baseline, and 70.6\% for DC-Merge, the best weight-only merge. The RegMean family stays at
60 to 61\%, since one image constrains few input directions (\cref{sec:metric}) and, unlike \method, it does not
complete them with the weights.

\begin{table}[H]
  \centering
  \footnotesize
  \setlength{\tabcolsep}{1.5pt}
  \begin{tabular}{@{}l c cccc c@{}}
    \toprule
    Method & \shortstack{Label-\\free} & KnOTS & $\mathrm{CE}_{0.014}$ & $\mathrm{CF}_{0.084}$ & $\mathrm{CF}_{0.188}$ & Avg. \\
    \midrule
    Zero-shot &  & 48.2 & 48.2 & 48.2 & 48.2 & 48.2 \\
    Individual experts &  & 84.0 & 88.4 & 89.5 & 89.0 & 87.7 \\
    \midrule
    \multicolumn{7}{@{}l}{\textit{Weight-only ($\lambda$ chosen with labels)}} \\
    TA~\cite{taskarithmetic} & \ding{55} & 51.4 & 69.3 & 76.1 & 69.2 & 66.5 \\
    TIES~\cite{ties} & \ding{55} & 51.4 & 68.9 & 74.9 & 68.2 & 65.9 \\
    KnOTS-TIES~\cite{knots} & \ding{55} & 56.1 & 71.3 & 74.6 & 69.2 & 67.8 \\
    TSV-M~\cite{tsv} & \ding{55} & 54.0 & \underline{72.7} & \underline{76.4} & \underline{72.1} & 68.8 \\
    Iso-C~\cite{isoc} & \ding{55} & 62.0 & 70.4 & 70.8 & 69.7 & 68.3 \\
    DC-Merge~\cite{dc_merge} & \ding{55} & 65.0 & \underline{72.7} & 73.3 & 71.4 & \underline{70.6} \\
    Core Space~\cite{corespace} & \ding{55} & 62.9 & 69.8 & 70.7 & 69.1 & 68.1 \\
    CT-Merging~\cite{ctmerging} & \ding{55} & \underline{66.6} & 69.3 & 69.4 & 67.0 & 68.1 \\
    ACTMat~\cite{actmat} & \ding{55} & 53.4 & 68.5 & 73.2 & 69.8 & 66.2 \\
    \midrule
    \multicolumn{7}{@{}l}{\textit{Activation-based (one unlabeled image per task)}} \\
    RegMean~\cite{regmean} & \checkmark & 52.2\std{0.9} & 66.1\std{0.6} & 64.2\std{1.5} & 59.0\std{0.2} & 60.4 \\
    RegMean++~\cite{regmeanpp} & \checkmark & 55.1\std{0.9} & 66.1\std{0.5} & 64.2\std{1.4} & 58.5\std{0.6} & 61.0 \\
    CrossMean~\cite{crossmean} & \checkmark & 53.4\std{1.6} & 66.2\std{0.5} & 63.7\std{1.5} & 58.1\std{0.3} & 60.3 \\
    AdaMerging~\cite{adamerging} & \checkmark & 57.6\std{0.7} & 66.1\std{1.7} & 73.8\std{1.4} & 67.6\std{0.9} & 66.3 \\
    TARA-B~\cite{tara} & \checkmark & 57.3\std{1.2} & 72.5\std{1.0} & 75.0\std{0.2} & 60.9\std{2.0} & 66.4 \\
    NSC~\cite{nsc} & \ding{55} & 50.0\std{1.0} & 69.5\std{0.6} & 75.0\std{0.7} & 69.6\std{0.4} & 66.0 \\
    \midrule
    \oursrow \textbf{\method} (ours) & \checkmark & \textbf{67.9}\std{0.7} & \textbf{76.0}\std{0.8} & \textbf{77.4}\std{0.9} & \textbf{73.9}\std{0.5} & \textbf{73.8} \\
    \bottomrule
  \end{tabular}
  \caption{\textbf{With one unlabeled image per task, \method outperforms every merge on each ViT-B/32 bank, including those that choose $\lambda$ on labeled images.} Test accuracy (\%) on eight tasks, mean $\pm$ std over five image draws. \emph{Label-free}: \ding{55}\ $\lambda$ chosen on 128 labeled images per task (\cref{sec:setup}). Bold: best; underlined: second. }
  \label{tab:vitb32}
\end{table}

\subsection{How far one image goes}
\label{sec:budget}
\begin{figure}[t]
    \centering
    \includegraphics[width=0.9\linewidth]{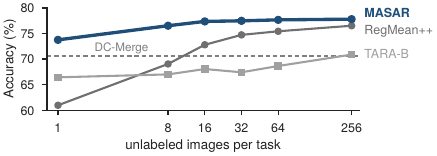}
    \caption{\textbf{One image goes further with \method.} Mean test accuracy (\%) on the four ViT-B/32 banks of
    \cref{tab:vitb32}. Dashed: DC-Merge, the best weight-only merge ($\lambda$ on labels).}
    \label{fig:budget}
\end{figure}

Since \method takes from the weights what one image misses, more images should help it less than merges that rely on
images alone. \Cref{fig:budget} confirms this against RegMean++ and TARA-B, which use no labels either: \method stays
first at every budget, while RegMean++ needs 32 images to pass \method with one (74.7
against 73.8\%).

\subsection{Transfer to ViT-L/14 and to a language model}
\label{sec:results-l14}

On ViT-L/14, \method chooses the same exponents as on ViT-B/32, although raw margins grow (\cref{sec:p-rule}).

\begin{table}[H]
  \centering
  \footnotesize
  \setlength{\tabcolsep}{1.1pt}
  \begin{tabular}{@{}l c ccc c c@{}}
    \toprule
    & & \multicolumn{4}{c}{ViT-L/14} & Llama-3-8B \\
    \cmidrule(lr){3-6}\cmidrule(l){7-7}
    Method & \shortstack{Label-\\free} & KnOTS & $\mathrm{CE}_{0.097}$ & $\mathrm{CF}_{0.231}$ & Avg. & NLI \\
    \midrule
    Zero-shot &  & 65.2 & 65.2 & 65.2 & 65.2 & 47.4 \\
    Individual experts &  & 92.4 & 93.9 & 93.4 & 93.3 & 92.5 \\
    \midrule
    \multicolumn{7}{@{}l}{\textit{Weight-only ($\lambda$ chosen with labels)}} \\
    TA~\cite{taskarithmetic} & \ding{55} & 68.5 & 87.2 & 84.7 & 80.1 & 83.4 \\
    KnOTS-TIES~\cite{knots} & \ding{55} & 72.5 & 86.9 & 84.9 & 81.5 & 85.6 \\
    TSV-M~\cite{tsv} & \ding{55} & 71.8 & \underline{88.5} & 86.8 & 82.4 & 85.7 \\
    Iso-C~\cite{isoc} & \ding{55} & 78.5 & 88.0 & 85.8 & 84.1 & 83.7 \\
    DC-Merge~\cite{dc_merge} & \ding{55} & 78.4 & \underline{88.5} & \underline{87.0} & \textbf{84.7} & \underline{85.8} \\
    Core Space~\cite{corespace} & \ding{55} & \underline{78.7} & 87.2 & 85.8 & 83.9 & 84.0 \\
    CT-Merging~\cite{ctmerging} & \ding{55} & \textbf{80.2} & 87.1 & 86.4 & 84.5 & 79.1 \\
    ACTMat~\cite{actmat} & \ding{55} & 70.7 & 88.2 & 85.9 & 81.6 & 85.3 \\
    \midrule
    \multicolumn{7}{@{}l}{\textit{Activation-based (one unlabeled input per task)}} \\
    RegMean~\cite{regmean} & \checkmark & 70.2\std{1.8} & 82.9\std{0.8} & 79.7\std{1.0} & 77.6 & 84.5\std{2.5} \\
    RegMean++~\cite{regmeanpp} & \checkmark & 71.1\std{1.6} & 83.1\std{0.7} & 79.7\std{1.0} & 78.0 & 85.1\std{2.3} \\
    CrossMean~\cite{crossmean} & \checkmark & 70.2\std{1.6} & 80.5\std{0.9} & 75.4\std{1.0} & 75.4 & 84.7\std{2.5} \\
    TARA-B~\cite{tara} & \checkmark & 78.5\std{1.1} & 87.2\std{0.2} & 83.5\std{0.2} & 83.1 & 75.0\std{3.9} \\
    NSC~\cite{nsc} & \ding{55} & 68.9\std{0.8} & 87.1\std{0.1} & 84.1\std{0.5} & 80.0 & 84.7\std{0.7} \\
    \midrule
    \oursrow \textbf{\method} (ours) & \checkmark & 76.0\std{1.7} & \textbf{89.6}\std{0.1} & \textbf{88.2}\std{0.1} & \underline{84.6} & \textbf{88.4}\std{0.5} \\
    \bottomrule
  \end{tabular}
  \caption{\textbf{On ViT-L/14, \method has the best label-free average and matches the best merge tuned on labels; on Llama-3-8B, it outperforms every merge.} Test accuracy (\%), mean $\pm$ std over five draws of one unlabeled input per task: an image for ViT-L/14, a sentence pair for NLI (Supp.~\ref{sec:supp-nli}). Notation as in \cref{tab:vitb32}. }
  \label{tab:vitl14}
\end{table}

Among label-free merges, \method has the highest average (\cref{tab:vitl14}), 84.6 against 83.1\% for TARA-B, and it
matches the best merges tuned on labels, DC-Merge (84.7\%) and CT-Merging (84.5\%). CT-Merging leads on KnOTS,
whose hesitant experts call for full flattening, but on the large-margin banks \method outperforms it and every other
merge. This highlights the margin-adaptive flattening of \method, which performs well whatever the confidence of
the experts (\cref{sec:banks}).

We also merge the six NLI LoRA experts that KnOTS trained on Llama-3-8B~\cite{knots,llama3}, each with its own
classification head: one unlabeled training pair per task replaces the image, the margin is the gap
between the two highest logits of the task head, and the rules are unchanged. With one pair per task, \method reaches 88.4\% (\cref{tab:vitl14}), 2.6 points above DC-Merge, the best merge that chooses $\lambda$ on labeled rows, and 3.3
above RegMean++. With the same rules, \method is also the best label-free merge of seven medical experts of the
vision-language model Gemma-4-31B~\cite{gemma4}, which mix image classification and visual question answering (Supp.~\ref{sec:supp-gemma}).

\subsection{What the label-free choices cost}
\label{sec:label-free}

\Cref{tab:label-free} separates the two choices that \method makes without labels: the exponent $p$ and the scales
$\lambda_t$.

\begin{table}[H]
  \centering
  \footnotesize
  \setlength{\tabcolsep}{3pt}
  \begin{tabular}{@{}ll c ccc@{}}
    \toprule
    Spectrum $p$ & Scales $\lambda_t$ & Label-free & ViT-B/32 & ViT-L/14 & NLI \\
    \midrule
    0 & $\lambda$, grid & \ding{55} & 72.0 & 83.9 & 83.8 \\
    grid & $\lambda$, grid & \ding{55} & 74.3 & 84.5 & 87.7 \\
    \refrow \cref{eq:p-rule} & $\lambda$, grid & \ding{55} & 74.4 & 84.5 & 87.6 \\
    \midrule
    min.\ entropy & $\lambda$, min.\ entropy & \checkmark & 64.7 & 78.8 & 79.9 \\
    \oursrow \cref{eq:p-rule} & $\frac12\kappa_t$ & \checkmark & 73.8 & 84.3 & 88.4 \\
    \bottomrule
  \end{tabular}
  \caption{\textbf{The relative margin chooses $p$ as well as labels do.} Mean test accuracy (\%) over the 9 ViT-B/32 and 5 ViT-L/14 banks and on the six NLI tasks of Llama-3-8B, five input draws. \emph{$\lambda$}: one scale for all experts. \emph{grid}: chosen on 128 labeled images per task. \emph{min.\ entropy}: chosen to minimize the prediction entropy, as in AdaMerging and TARA. Gray: the labeled reference.}
  \label{tab:label-free}
\end{table}

The margin chooses the exponent as well as labels do. With one $\lambda$ chosen on labels in the first three rows, the
$p$ of \cref{eq:p-rule} matches a search of $p$ on labels on all three models. Full flattening loses 0.6 to 3.9
points, most on NLI, where every expert is confident, whereas the ViT averages also include small-margin banks that
full flattening suits. Dropping labels for the scales also costs little, since $\frac12\kappa_t$ (last row) stays
within 0.7 points of the labeled $\lambda$. The entropy criterion of AdaMerging and TARA falls 5 to 9 points below
\method, because it rewards confident predictions rather than correct ones.

\section{Conclusion}
\label{sec:conclusion}

How LoRA experts should be merged depends on how they were trained. Yet existing merges treat every bank the same
way, so the merge that works best on one bank can fall behind on another. We
introduced \method for merging when data are scarce, from a single unlabeled image per task. It reads the confidence of
the experts from their margins on this image, sets from it how much of their spectra to keep, and completes what the image
misses with the weights, so that each module is merged in closed form, without labels or gradient steps.
\method is thus the best merge on 11 of our 14 banks, from the most hesitant experts to the most confident ones, and
on every ViT-B/32 bank of \cref{tab:vitb32} it outperforms even the merges tuned on labels. The same rules carry over
unchanged to ViT-L/14 and to a language model, where \method is the best label-free merge.

\paragraph{Limitations.} The rule linking margins to the exponent is set on KnOTS and on banks that differ only in their loss, all of rank 16; other ranks and training budgets remain untested. Banks mixing training recipes call for expert-specific exponents, which gain up to 3 points (Supp.~\ref{sec:supp-mixed}). Finally, \method remains untested on long free-form generation, beyond short answers.

\section*{LLM Usage Statement}
Large language models (Claude by Anthropic and ChatGPT by OpenAI) assisted with the experiment code, the analysis of the results, the literature search and the
writing and figure layout. The authors reviewed all content and take full responsibility for it.

{
    \small
    \bibliographystyle{ieeenat_fullname}
    \bibliography{main}

\begin{thebibliography}{79}
\providecommand{\natexlab}[1]{#1}
\providecommand{\url}[1]{\texttt{#1}}
\expandafter\ifx\csname urlstyle\endcsname\relax
  \providecommand{\doi}[1]{doi: #1}\else
  \providecommand{\doi}{doi: \begingroup \urlstyle{rm}\Url}\fi

\bibitem[Antol et~al.(2015)Antol, Agrawal, Lu, Mitchell, Batra, Zitnick, and
  Parikh]{vqa}
Stanislaw Antol, Aishwarya Agrawal, Jiasen Lu, Margaret Mitchell, Dhruv Batra,
  C.~Lawrence Zitnick, and Devi Parikh.
\newblock {VQA}: Visual question answering.
\newblock In \emph{Proceedings of the IEEE International Conference on Computer
  Vision}, 2015.

\bibitem[Ben~Abacha et~al.(2019)Ben~Abacha, Hasan, Datla, Liu, Demner-Fushman,
  and M{\"u}ller]{vqamed}
Asma Ben~Abacha, Sadid~A. Hasan, Vivek~V. Datla, Joey Liu, Dina Demner-Fushman,
  and Henning M{\"u}ller.
\newblock {VQA-Med}: Overview of the medical visual question answering task at
  {ImageCLEF} 2019.
\newblock In \emph{Working Notes of {CLEF} 2019 -- Conference and Labs of the
  Evaluation Forum}, 2019.

\bibitem[Bossard et~al.(2014)Bossard, Guillaumin, and Van~Gool]{food101}
Lukas Bossard, Matthieu Guillaumin, and Luc Van~Gool.
\newblock Food-101 -- mining discriminative components with random forests.
\newblock In \emph{European Conference on Computer Vision}, 2014.

\bibitem[Bowman et~al.(2015)Bowman, Angeli, Potts, and Manning]{snli}
Samuel~R. Bowman, Gabor Angeli, Christopher Potts, and Christopher~D. Manning.
\newblock A large annotated corpus for learning natural language inference.
\newblock In \emph{Proceedings of the Conference on Empirical Methods in
  Natural Language Processing}, pages 632--642, 2015.

\bibitem[Buzzega et~al.(2026)Buzzega, Salami, Porrello, and
  Calderara]{chainofmerges}
Pietro Buzzega, Riccardo Salami, Angelo Porrello, and Simone Calderara.
\newblock Rethinking layer-wise model merging through chain of merges.
\newblock In \emph{Advances in Neural Information Processing Systems}, 2026.

\bibitem[Chen et~al.(2025)Chen, Wang, Li, Zhu, and Chen]{iteris}
Hongxu Chen, Zhen Wang, Runshi Li, Bowei Zhu, and Long Chen.
\newblock {IterIS}: Iterative inference-solving alignment for {LoRA} merging.
\newblock In \emph{Proceedings of the IEEE/CVF Conference on Computer Vision
  and Pattern Recognition}, pages 4829--4838, 2025.

\bibitem[Cheng et~al.(2017)Cheng, Han, and Lu]{resisc45}
Gong Cheng, Junwei Han, and Xiaoqiang Lu.
\newblock Remote sensing image scene classification: Benchmark and state of the
  art.
\newblock \emph{Proceedings of the IEEE}, 105\penalty0 (10):\penalty0
  1865--1883, 2017.

\bibitem[Cheng et~al.(2025)Cheng, Xiong, Wei, Zhu, and Yuan]{wudi}
Runxi Cheng, Feng Xiong, Yongxian Wei, Wanyun Zhu, and Chun Yuan.
\newblock Whoever started the interference should end it: Guiding data-free
  model merging via task vectors.
\newblock In \emph{International Conference on Machine Learning}, pages
  10121--10143, 2025.

\bibitem[Cimpoi et~al.(2014)Cimpoi, Maji, Kokkinos, Mohamed, and Vedaldi]{dtd}
Mircea Cimpoi, Subhransu Maji, Iasonas Kokkinos, Sammy Mohamed, and Andrea
  Vedaldi.
\newblock Describing textures in the wild.
\newblock In \emph{Proceedings of the IEEE Conference on Computer Vision and
  Pattern Recognition}, pages 3606--3613, 2014.

\bibitem[Coates et~al.(2011)Coates, Ng, and Lee]{stl10}
Adam Coates, Andrew~Y. Ng, and Honglak Lee.
\newblock An analysis of single-layer networks in unsupervised feature
  learning.
\newblock In \emph{Proceedings of the International Conference on Artificial
  Intelligence and Statistics}, 2011.

\bibitem[Cobbe et~al.(2021)Cobbe, Kosaraju, Bavarian, Chen, Jun, Kaiser,
  Plappert, Tworek, Hilton, Nakano, Hesse, and Schulman]{gsm8k}
Karl Cobbe, Vineet Kosaraju, Mohammad Bavarian, Mark Chen, Heewoo Jun, Lukasz
  Kaiser, Matthias Plappert, Jerry Tworek, Jacob Hilton, Reiichiro Nakano,
  Christopher Hesse, and John Schulman.
\newblock Training verifiers to solve math word problems.
\newblock \emph{arXiv preprint arXiv:2110.14168}, 2021.

\bibitem[Deng et~al.(2009)Deng, Dong, Socher, Li, Li, and Fei-Fei]{imagenet}
Jia Deng, Wei Dong, Richard Socher, Li-Jia Li, Kai Li, and Li Fei-Fei.
\newblock {ImageNet}: A large-scale hierarchical image database.
\newblock In \emph{Proceedings of the IEEE Conference on Computer Vision and
  Pattern Recognition}, 2009.

\bibitem[Dosovitskiy et~al.(2021)Dosovitskiy, Beyer, Kolesnikov, Weissenborn,
  Zhai, Unterthiner, Dehghani, Minderer, Heigold, Gelly, Uszkoreit, and
  Houlsby]{vit}
Alexey Dosovitskiy, Lucas Beyer, Alexander Kolesnikov, Dirk Weissenborn,
  Xiaohua Zhai, Thomas Unterthiner, Mostafa Dehghani, Matthias Minderer, Georg
  Heigold, Sylvain Gelly, Jakob Uszkoreit, and Neil Houlsby.
\newblock An image is worth 16x16 words: Transformers for image recognition at
  scale.
\newblock In \emph{International Conference on Learning Representations}, 2021.

\bibitem[Eliasof(2026)]{crossmean}
Moshe Eliasof.
\newblock {CrossMean}: Model merging by pairing merged and expert activations.
\newblock \emph{arXiv preprint arXiv:2610.14762}, 2026.

\bibitem[Fan and Hoffman(1955)]{fanhoffman}
Ky Fan and A.~J. Hoffman.
\newblock Some metric inequalities in the space of matrices.
\newblock \emph{Proceedings of the American Mathematical Society}, 6\penalty0
  (1):\penalty0 111--116, 1955.

\bibitem[Gargiulo et~al.(2025)Gargiulo, Crisostomi, Bucarelli, Scardapane,
  Silvestri, and Rodol{\`a}]{tsv}
Antonio~Andrea Gargiulo, Donato Crisostomi, Maria~Sofia Bucarelli, Simone
  Scardapane, Fabrizio Silvestri, and Emanuele Rodol{\`a}.
\newblock Task singular vectors: Reducing task interference in model merging.
\newblock In \emph{Proceedings of the IEEE/CVF Conference on Computer Vision
  and Pattern Recognition}, pages 18695--18705, 2025.

\bibitem[{Gemma Team}(2026)]{gemma4}
{Gemma Team}.
\newblock {Gemma} 4 technical report.
\newblock \emph{arXiv preprint arXiv:2607.02770}, 2026.

\bibitem[Goodfellow et~al.(2013)Goodfellow, Erhan, Carrier, Courville, Mirza,
  Hamner, Cukierski, Tang, Thaler, Lee, et~al.]{fer2013}
Ian~J. Goodfellow, Dumitru Erhan, Pierre~Luc Carrier, Aaron Courville, Mehdi
  Mirza, Ben Hamner, Will Cukierski, Yichuan Tang, David Thaler, Dong-Hyun Lee,
  et~al.
\newblock Challenges in representation learning: A report on three machine
  learning contests.
\newblock In \emph{International Conference on Neural Information Processing},
  2013.

\bibitem[Goyal et~al.(2017)Goyal, Khot, Summers-Stay, Batra, and Parikh]{vqav2}
Yash Goyal, Tejas Khot, Douglas Summers-Stay, Dhruv Batra, and Devi Parikh.
\newblock Making the {V} in {VQA} matter: Elevating the role of image
  understanding in visual question answering.
\newblock In \emph{Proceedings of the IEEE Conference on Computer Vision and
  Pattern Recognition}, 2017.

\bibitem[Grattafiori et~al.(2024)Grattafiori, Dubey, Jauhri, Pandey, Kadian,
  et~al.]{llama3}
Aaron Grattafiori, Abhimanyu Dubey, Abhinav Jauhri, Abhinav Pandey, Abhishek
  Kadian, et~al.
\newblock The {Llama} 3 herd of models.
\newblock \emph{arXiv preprint arXiv:2407.21783}, 2024.

\bibitem[Gu et~al.(2026)Gu, Cai, Wang, Wang, Wang, Wang, Huang, Lu, Wu, and
  Yang]{featcal}
Yanggan Gu, Shuo Cai, Zihao Wang, Wenjun Wang, Yuanyi Wang, Pengkai Wang, Sirui
  Huang, Su Lu, Jianmin Wu, and Hongxia Yang.
\newblock {FeatCal}: Feature calibration for post-merging models.
\newblock \emph{arXiv preprint arXiv:2605.13030}, 2026.

\bibitem[Gurari et~al.(2018)Gurari, Li, Stangl, Guo, Lin, Grauman, Luo, and
  Bigham]{vizwiz}
Danna Gurari, Qing Li, Abigale~J. Stangl, Anhong Guo, Chi Lin, Kristen Grauman,
  Jiebo Luo, and Jeffrey~P. Bigham.
\newblock {VizWiz} grand challenge: Answering visual questions from blind
  people.
\newblock In \emph{Proceedings of the IEEE Conference on Computer Vision and
  Pattern Recognition}, 2018.

\bibitem[Hameed et~al.(2026)Hameed, Tam, Notsawo, Raffel, and
  Rabusseau]{actmat}
Marawan Gamal~Abdel Hameed, Derek Tam, Pascal Jr~Tikeng Notsawo, Colin Raffel,
  and Guillaume Rabusseau.
\newblock Model merging via data-free covariance estimation.
\newblock \emph{arXiv preprint arXiv:2604.01329}, 2026.

\bibitem[He et~al.(2020)He, Zhang, Mou, Xing, and Xie]{pathvqa}
Xuehai He, Yichen Zhang, Luntian Mou, Eric Xing, and Pengtao Xie.
\newblock {PathVQA}: 30000+ questions for medical visual question answering.
\newblock \emph{arXiv preprint arXiv:2003.10286}, 2020.

\bibitem[Helber et~al.(2019)Helber, Bischke, Dengel, and Borth]{eurosat}
Patrick Helber, Benjamin Bischke, Andreas Dengel, and Damian Borth.
\newblock {EuroSAT}: A novel dataset and deep learning benchmark for land use
  and land cover classification.
\newblock \emph{IEEE Journal of Selected Topics in Applied Earth Observations
  and Remote Sensing}, 12\penalty0 (7):\penalty0 2217--2226, 2019.

\bibitem[Higham(1986)]{higham}
Nicholas~J. Higham.
\newblock Computing the polar decomposition---with applications.
\newblock \emph{SIAM Journal on Scientific and Statistical Computing},
  7\penalty0 (4):\penalty0 1160--1174, 1986.

\bibitem[Horoi et~al.(2026)Horoi, Wolf, Belilovsky, and
  Dziugaite]{overtraining}
Stefan Horoi, Guy Wolf, Eugene Belilovsky, and Gintare~Karolina Dziugaite.
\newblock From memorization to parameter interference: How overtraining experts
  harms model merging.
\newblock In \emph{International Conference on Machine Learning}, pages
  43896--43918, 2026.

\bibitem[Hu et~al.(2022)Hu, Shen, Wallis, Allen-Zhu, Li, Wang, Wang, and
  Chen]{lora}
Edward~J. Hu, Yelong Shen, Phillip Wallis, Zeyuan Allen-Zhu, Yuanzhi Li, Shean
  Wang, Lu Wang, and Weizhu Chen.
\newblock {LoRA}: Low-rank adaptation of large language models.
\newblock In \emph{International Conference on Learning Representations}, 2022.

\bibitem[Huang et~al.(2024)Huang, Liu, Lin, Pang, Du, and Lin]{lorahub}
Chengsong Huang, Qian Liu, Bill~Yuchen Lin, Tianyu Pang, Chao Du, and Min Lin.
\newblock {LoraHub}: Efficient cross-task generalization via dynamic {LoRA}
  composition.
\newblock In \emph{Conference on Language Modeling}, 2024.

\bibitem[Ilharco et~al.(2023)Ilharco, Ribeiro, Wortsman, Gururangan, Schmidt,
  Hajishirzi, and Farhadi]{taskarithmetic}
Gabriel Ilharco, Marco~Tulio Ribeiro, Mitchell Wortsman, Suchin Gururangan,
  Ludwig Schmidt, Hannaneh Hajishirzi, and Ali Farhadi.
\newblock Editing models with task arithmetic.
\newblock In \emph{International Conference on Learning Representations}, 2023.

\bibitem[Jeong et~al.(2026)Jeong, Lee, and Yoon]{tara}
Wooseong Jeong, Wonyoung Lee, and Kuk-Jin Yoon.
\newblock Preference-aligned {LoRA} merging: Preserving subspace coverage and
  addressing directional anisotropy.
\newblock In \emph{Proceedings of the IEEE/CVF Conference on Computer Vision
  and Pattern Recognition}, 2026.

\bibitem[Jin et~al.(2023)Jin, Ren, Preotiuc-Pietro, and Cheng]{regmean}
Xisen Jin, Xiang Ren, Daniel Preotiuc-Pietro, and Pengxiang Cheng.
\newblock Dataless knowledge fusion by merging weights of language models.
\newblock In \emph{International Conference on Learning Representations}, 2023.

\bibitem[Khot et~al.(2018)Khot, Sabharwal, and Clark]{scitail}
Tushar Khot, Ashish Sabharwal, and Peter Clark.
\newblock {SciTaiL}: A textual entailment dataset from science question
  answering.
\newblock In \emph{Proceedings of the AAAI Conference on Artificial
  Intelligence}, pages 5189--5197, 2018.

\bibitem[Kingma and Ba(2015)]{adam}
Diederik~P. Kingma and Jimmy Ba.
\newblock Adam: A method for stochastic optimization.
\newblock In \emph{International Conference on Learning Representations}, 2015.

\bibitem[Krause et~al.(2013)Krause, Stark, Deng, and Fei-Fei]{cars}
Jonathan Krause, Michael Stark, Jia Deng, and Li Fei-Fei.
\newblock {3D} object representations for fine-grained categorization.
\newblock In \emph{Proceedings of the IEEE International Conference on Computer
  Vision Workshops}, pages 554--561, 2013.

\bibitem[Krizhevsky(2009)]{cifar100}
Alex Krizhevsky.
\newblock Learning multiple layers of features from tiny images.
\newblock Technical report, University of Toronto, 2009.

\bibitem[LeCun et~al.(1998)LeCun, Bottou, Bengio, and Haffner]{mnist}
Yann LeCun, L{\'e}on Bottou, Yoshua Bengio, and Patrick Haffner.
\newblock Gradient-based learning applied to document recognition.
\newblock \emph{Proceedings of the IEEE}, 86\penalty0 (11):\penalty0
  2278--2324, 1998.

\bibitem[Lee et~al.(2026)Lee, Jeong, and Yoon]{nsc}
Wonyoung Lee, Wooseong Jeong, and Kuk-Jin Yoon.
\newblock Label-free cross-task {LoRA} merging with null-space compression.
\newblock In \emph{Proceedings of the IEEE/CVF Conference on Computer Vision
  and Pattern Recognition}, pages 847--859, 2026.

\bibitem[Li et~al.(2026)Li, Han, Su, and Ji]{reforge}
Kaiyang Li, Shaobo Han, Qing Su, and Shihao Ji.
\newblock {ReForge}: Refining merged models with anchor-regularized regression.
\newblock \emph{arXiv preprint arXiv:2605.12843}, 2026.

\bibitem[Li et~al.(2019{\natexlab{a}})Li, Xu, Wang, Jiang, and Liu]{lag}
Liu Li, Mai Xu, Xiaofei Wang, Lai Jiang, and Hanruo Liu.
\newblock Attention based glaucoma detection: A large-scale database and {CNN}
  model.
\newblock In \emph{Proceedings of the IEEE/CVF Conference on Computer Vision
  and Pattern Recognition}, pages 10571--10580, 2019{\natexlab{a}}.

\bibitem[Li et~al.(2019{\natexlab{b}})Li, Gao, Wang, Guo, Liu, and Kang]{ddr}
Tao Li, Yingqi Gao, Kai Wang, Song Guo, Hanruo Liu, and Hong Kang.
\newblock Diagnostic assessment of deep learning algorithms for diabetic
  retinopathy screening.
\newblock \emph{Information Sciences}, 501:\penalty0 511--522,
  2019{\natexlab{b}}.

\bibitem[Liu et~al.(2021)Liu, Zhan, Xu, Ma, Yang, and Wu]{slake}
Bo Liu, Li-Ming Zhan, Li Xu, Lin Ma, Yan Yang, and Xiao-Ming Wu.
\newblock {SLAKE}: A semantically-labeled knowledge-enhanced dataset for
  medical visual question answering.
\newblock In \emph{IEEE International Symposium on Biomedical Imaging}, pages
  1650--1654, 2021.

\bibitem[Liu et~al.(2024)Liu, Li, Li, and Lee]{llava15}
Haotian Liu, Chunyuan Li, Yuheng Li, and Yong~Jae Lee.
\newblock Improved baselines with visual instruction tuning.
\newblock In \emph{Proceedings of the IEEE/CVF Conference on Computer Vision
  and Pattern Recognition}, 2024.

\bibitem[Liu et~al.(2020)Liu, Wu, Ban, Wang, and Cheng]{tbx11k}
Yun Liu, Yu-Huan Wu, Yunfeng Ban, Huifang Wang, and Ming-Ming Cheng.
\newblock Rethinking computer-aided tuberculosis diagnosis.
\newblock In \emph{Proceedings of the IEEE/CVF Conference on Computer Vision
  and Pattern Recognition}, pages 2646--2655, 2020.

\bibitem[Loshchilov and Hutter(2017)]{sgdr}
Ilya Loshchilov and Frank Hutter.
\newblock {SGDR}: Stochastic gradient descent with warm restarts.
\newblock In \emph{International Conference on Learning Representations}, 2017.

\bibitem[Loshchilov and Hutter(2019)]{adamw}
Ilya Loshchilov and Frank Hutter.
\newblock Decoupled weight decay regularization.
\newblock In \emph{International Conference on Learning Representations}, 2019.

\bibitem[Lu et~al.(2021)Lu, Qiu, Chen, Xia, Zhao, Zhang, Yu, Liang, and
  Zhu]{iconqa}
Pan Lu, Liang Qiu, Jiaqi Chen, Tony Xia, Yizhou Zhao, Wei Zhang, Zhou Yu,
  Xiaodan Liang, and Song-Chun Zhu.
\newblock {IconQA}: A new benchmark for abstract diagram understanding and
  visual language reasoning.
\newblock In \emph{Proceedings of the Neural Information Processing Systems
  Track on Datasets and Benchmarks}, 2021.

\bibitem[Lu et~al.(2022)Lu, Mishra, Xia, Qiu, Chang, Zhu, Tafjord, Clark, and
  Kalyan]{scienceqa}
Pan Lu, Swaroop Mishra, Tony Xia, Liang Qiu, Kai-Wei Chang, Song-Chun Zhu,
  Oyvind Tafjord, Peter Clark, and Ashwin Kalyan.
\newblock Learn to explain: Multimodal reasoning via thought chains for science
  question answering.
\newblock In \emph{Advances in Neural Information Processing Systems}, 2022.

\bibitem[Marczak et~al.(2025)Marczak, Magistri, Cygert, Twardowski, Bagdanov,
  and van~de Weijer]{isoc}
Daniel Marczak, Simone Magistri, Sebastian Cygert, Bart{\l}omiej Twardowski,
  Andrew~D. Bagdanov, and Joost van~de Weijer.
\newblock No task left behind: Isotropic model merging with common and
  task-specific subspaces.
\newblock In \emph{International Conference on Machine Learning}, pages
  43177--43199, 2025.

\bibitem[Marelli et~al.(2014)Marelli, Menini, Baroni, Bentivogli, Bernardi, and
  Zamparelli]{sick}
Marco Marelli, Stefano Menini, Marco Baroni, Luisa Bentivogli, Raffaella
  Bernardi, and Roberto Zamparelli.
\newblock A {SICK} cure for the evaluation of compositional distributional
  semantic models.
\newblock In \emph{Proceedings of the International Conference on Language
  Resources and Evaluation}, pages 216--223, 2014.

\bibitem[Matena and Raffel(2022)]{fisher}
Michael Matena and Colin Raffel.
\newblock Merging models with {Fisher}-weighted averaging.
\newblock In \emph{Advances in Neural Information Processing Systems}, pages
  17703--17716, 2022.

\bibitem[Mishra et~al.(2019)Mishra, Shekhar, Singh, and Chakraborty]{ocrvqa}
Anand Mishra, Shashank Shekhar, Ajeet~Kumar Singh, and Anirban Chakraborty.
\newblock {OCR-VQA}: Visual question answering by reading text in images.
\newblock In \emph{International Conference on Document Analysis and
  Recognition}, 2019.

\bibitem[Netzer et~al.(2011)Netzer, Wang, Coates, Bissacco, Wu, and Ng]{svhn}
Yuval Netzer, Tao Wang, Adam Coates, Alessandro Bissacco, Bo Wu, and Andrew~Y.
  Ng.
\newblock Reading digits in natural images with unsupervised feature learning.
\newblock In \emph{NIPS Workshop on Deep Learning and Unsupervised Feature
  Learning}, 2011.

\bibitem[Nguyen et~al.(2026)Nguyen, {Dang Huu-Tien}, Suzuki, and
  Nguyen]{regmeanpp}
The-Hai Nguyen, {Dang Huu-Tien}, Takeshi Suzuki, and Le-Minh Nguyen.
\newblock {RegMean++}: Enhancing effectiveness and generalization of regression
  mean for model merging.
\newblock \emph{Transactions on Machine Learning Research}, 2026.

\bibitem[Nilsback and Zisserman(2008)]{flowers102}
Maria-Elena Nilsback and Andrew Zisserman.
\newblock Automated flower classification over a large number of classes.
\newblock In \emph{Indian Conference on Computer Vision, Graphics and Image
  Processing}, 2008.

\bibitem[Ortiz-Jimenez et~al.(2023)Ortiz-Jimenez, Favero, and
  Frossard]{tangent}
Guillermo Ortiz-Jimenez, Alessandro Favero, and Pascal Frossard.
\newblock Task arithmetic in the tangent space: Improved editing of pre-trained
  models.
\newblock In \emph{Advances in Neural Information Processing Systems}, pages
  66727--66754, 2023.

\bibitem[Panariello et~al.(2025)Panariello, Marczak, Magistri, Porrello,
  Twardowski, Bagdanov, Calderara, and van~de Weijer]{corespace}
Aniello Panariello, Daniel Marczak, Simone Magistri, Angelo Porrello,
  Bart{\l}omiej Twardowski, Andrew~D. Bagdanov, Simone Calderara, and Joost
  van~de Weijer.
\newblock Accurate and efficient low-rank model merging in core space.
\newblock In \emph{Advances in Neural Information Processing Systems}, pages
  61793--61825, 2025.

\bibitem[Parkhi et~al.(2012)Parkhi, Vedaldi, Zisserman, and Jawahar]{pets}
Omkar~M. Parkhi, Andrea Vedaldi, Andrew Zisserman, and C.~V. Jawahar.
\newblock Cats and dogs.
\newblock In \emph{Proceedings of the IEEE Conference on Computer Vision and
  Pattern Recognition}, 2012.

\bibitem[Radford et~al.(2021)Radford, Kim, Hallacy, Ramesh, Goh, Agarwal,
  Sastry, Askell, Mishkin, Clark, Krueger, and Sutskever]{clip}
Alec Radford, Jong~Wook Kim, Chris Hallacy, Aditya Ramesh, Gabriel Goh,
  Sandhini Agarwal, Girish Sastry, Amanda Askell, Pamela Mishkin, Jack Clark,
  Gretchen Krueger, and Ilya Sutskever.
\newblock Learning transferable visual models from natural language
  supervision.
\newblock In \emph{International Conference on Machine Learning}, pages
  8748--8763, 2021.

\bibitem[Ryum and Kang(2026)]{ctmerging}
Keumseo Ryum and Joonhyuk Kang.
\newblock {CT-Merging}: Consensus directions and task-specific scaling for
  {LoRA} adapter merging.
\newblock \emph{arXiv preprint arXiv:2607.20561}, 2026.

\bibitem[Spanhol et~al.(2016)Spanhol, Oliveira, Petitjean, and
  Heutte]{breakhis}
Fabio~A. Spanhol, Luiz~S. Oliveira, Caroline Petitjean, and Laurent Heutte.
\newblock A dataset for breast cancer histopathological image classification.
\newblock \emph{IEEE Transactions on Biomedical Engineering}, 63\penalty0
  (7):\penalty0 1455--1462, 2016.

\bibitem[Stallkamp et~al.(2011)Stallkamp, Schlipsing, Salmen, and Igel]{gtsrb}
Johannes Stallkamp, Marc Schlipsing, Jan Salmen, and Christian Igel.
\newblock The {German} traffic sign recognition benchmark: A multi-class
  classification competition.
\newblock In \emph{International Joint Conference on Neural Networks}, pages
  1453--1460, 2011.

\bibitem[Stoica et~al.(2025)Stoica, Ramesh, Ecsedi, Choshen, and
  Hoffman]{knots}
George Stoica, Pratik Ramesh, Boglarka Ecsedi, Leshem Choshen, and Judy
  Hoffman.
\newblock Model merging with {SVD} to tie the {Knots}.
\newblock In \emph{International Conference on Learning Representations}, 2025.

\bibitem[Wang et~al.(2019)Wang, Singh, Michael, Hill, Levy, and Bowman]{glue}
Alex Wang, Amanpreet Singh, Julian Michael, Felix Hill, Omer Levy, and
  Samuel~R. Bowman.
\newblock {GLUE}: A multi-task benchmark and analysis platform for natural
  language understanding.
\newblock In \emph{International Conference on Learning Representations}, 2019.

\bibitem[Wang et~al.(2018)Wang, Wang, Zhou, Ji, Gong, Zhou, Li, and
  Liu]{cosface}
Hao Wang, Yitong Wang, Zheng Zhou, Xing Ji, Dihong Gong, Jingchao Zhou, Zhifeng
  Li, and Wei Liu.
\newblock {CosFace}: Large margin cosine loss for deep face recognition.
\newblock In \emph{Proceedings of the IEEE Conference on Computer Vision and
  Pattern Recognition}, pages 5265--5274, 2018.

\bibitem[Williams et~al.(2018)Williams, Nangia, and Bowman]{mnli}
Adina Williams, Nikita Nangia, and Samuel~R. Bowman.
\newblock A broad-coverage challenge corpus for sentence understanding through
  inference.
\newblock In \emph{Proceedings of the Conference of the North {American}
  Chapter of the Association for Computational Linguistics: Human Language
  Technologies}, pages 1112--1122, 2018.

\bibitem[Wortsman et~al.(2022)Wortsman, Ilharco, Gadre, Roelofs, Gontijo-Lopes,
  Morcos, Namkoong, Farhadi, Carmon, Kornblith, and Schmidt]{modelsoups}
Mitchell Wortsman, Gabriel Ilharco, Samir~Yitzhak Gadre, Rebecca Roelofs,
  Raphael Gontijo-Lopes, Ari~S. Morcos, Hongseok Namkoong, Ali Farhadi, Yair
  Carmon, Simon Kornblith, and Ludwig Schmidt.
\newblock Model soups: Averaging weights of multiple fine-tuned models improves
  accuracy without increasing inference time.
\newblock In \emph{International Conference on Machine Learning}, pages
  23965--23998, 2022.

\bibitem[Xiao et~al.(2017)Xiao, Rasul, and Vollgraf]{fashionmnist}
Han Xiao, Kashif Rasul, and Roland Vollgraf.
\newblock {Fashion-MNIST}: a novel image dataset for benchmarking machine
  learning algorithms.
\newblock \emph{arXiv preprint arXiv:1708.07747}, 2017.

\bibitem[Xiao et~al.(2010)Xiao, Hays, Ehinger, Oliva, and Torralba]{sun397}
Jianxiong Xiao, James Hays, Krista~A. Ehinger, Aude Oliva, and Antonio
  Torralba.
\newblock {SUN} database: Large-scale scene recognition from abbey to zoo.
\newblock In \emph{Proceedings of the IEEE Conference on Computer Vision and
  Pattern Recognition}, pages 3485--3492, 2010.

\bibitem[Xu et~al.(2026{\natexlab{a}})Xu, Wu, Lin, Huang, Zhu, Shu, and
  Qin]{acemerging}
Bo Xu, Haotian Wu, Hehai Lin, Weiquan Huang, Beier Zhu, Yao Shu, and Chengwei
  Qin.
\newblock {ACE-Merging}: Data-free model merging with adaptive covariance
  estimation.
\newblock In \emph{Proceedings of the IEEE/CVF Conference on Computer Vision
  and Pattern Recognition}, pages 29472--29482, 2026{\natexlab{a}}.

\bibitem[Xu et~al.(2026{\natexlab{b}})Xu, Min, MacDonald, Tarmoun, Mallada, Su,
  and Vidal]{malora}
Ziqing Xu, Hancheng Min, Lachlan~Ewen MacDonald, Salma Tarmoun, Enrique
  Mallada, Weijie Su, and Ren{\'e} Vidal.
\newblock A margin perspective on {LoRA}: Robustness to catastrophic forgetting
  and adapter merging ({MaLoRA}).
\newblock In \emph{Advances in Neural Information Processing Systems},
  2026{\natexlab{b}}.

\bibitem[Yadav et~al.(2023)Yadav, Tam, Choshen, Raffel, and Bansal]{ties}
Prateek Yadav, Derek Tam, Leshem Choshen, Colin Raffel, and Mohit Bansal.
\newblock {TIES-Merging}: Resolving interference when merging models.
\newblock In \emph{Advances in Neural Information Processing Systems}, pages
  7093--7115, 2023.

\bibitem[Yang et~al.(2024)Yang, Wang, Shen, Liu, Guo, Wang, and
  Tao]{adamerging}
Enneng Yang, Zhenyi Wang, Li Shen, Shiwei Liu, Guibing Guo, Xingwei Wang, and
  Dacheng Tao.
\newblock {AdaMerging}: Adaptive model merging for multi-task learning.
\newblock In \emph{International Conference on Learning Representations}, 2024.

\bibitem[Yoshida et~al.(2026)Yoshida, Naraki, Horie, Shimizu, Mitliagkas, and
  Naganuma]{distac}
Kotaro Yoshida, Yuji Naraki, Takafumi Horie, Ryotaro Shimizu, Ioannis
  Mitliagkas, and Hiroki Naganuma.
\newblock {DisTaC}: Conditioning task vectors via distillation for robust model
  merging.
\newblock In \emph{International Conference on Learning Representations}, 2026.

\bibitem[Young et~al.(2014)Young, Lai, Hodosh, and Hockenmaier]{flickr30k}
Peter Young, Alice Lai, Micah Hodosh, and Julia Hockenmaier.
\newblock From image descriptions to visual denotations: New similarity metrics
  for semantic inference over event descriptions.
\newblock \emph{Transactions of the Association for Computational Linguistics},
  2:\penalty0 67--78, 2014.

\bibitem[Yu et~al.(2016)Yu, Poirson, Yang, Berg, and Berg]{refcoco}
Licheng Yu, Patrick Poirson, Shan Yang, Alexander~C. Berg, and Tamara~L. Berg.
\newblock Modeling context in referring expressions.
\newblock In \emph{European Conference on Computer Vision}, 2016.

\bibitem[Zeng et~al.(2025)Zeng, Guo, Zhu, Shen, and Tang]{robustmerge}
Fanhu Zeng, Haiyang Guo, Fei Zhu, Li Shen, and Hao Tang.
\newblock {RobustMerge}: Parameter-efficient model merging for {MLLMs} with
  direction robustness.
\newblock In \emph{Advances in Neural Information Processing Systems}, 2025.

\bibitem[Zhang and Zhou(2025)]{osrm}
Haobo Zhang and Jiayu Zhou.
\newblock Unraveling {LoRA} interference: Orthogonal subspaces for robust model
  merging.
\newblock In \emph{Proceedings of the Annual Meeting of the Association for
  Computational Linguistics}, pages 26459--26472, 2025.

\bibitem[Zhang et~al.(2026)Zhang, Zhou, Luo, Di, Zhang, and Wei]{dc_merge}
Han-Chen Zhang, Zi-Hao Zhou, Mao-Lin Luo, Shimin Di, Min-Ling Zhang, and Tong
  Wei.
\newblock {DC-Merge}: Improving model merging with directional consistency.
\newblock In \emph{Proceedings of the IEEE/CVF Conference on Computer Vision
  and Pattern Recognition}, pages 22248--22258, 2026.

\end{thebibliography}
}

\clearpage
\maketitlesupplementary
\setcounter{section}{0}
\renewcommand{\thesection}{\Alph{section}}
\renewcommand{\theHsection}{supp.\Alph{section}}
\newcommand{\splitname}[1]{\textit{#1}}
\setcounter{dbltopnumber}{3}
\renewcommand{\dbltopfraction}{0.85}
\renewcommand{\dblfloatpagefraction}{0.8}

\section{Reconstruction, scaling and margins}
\label{sec:supp-math}
\Cref{sec:reconstruction,sec:margins} show what each step of \method does. This section gives the short arguments
behind these results, for one module $\ell$ of expert $t$, whose indices we drop when clear. \Cref{tab:supp-notation}
lists the notation.

\begin{table}[H]
    \centering
    \footnotesize
    \setlength{\tabcolsep}{3pt}
    \begin{tabular}{@{}l >{\raggedright\arraybackslash}p{6cm}@{}}
        \toprule
        Symbol & Meaning \\
        \midrule
        $T$, $L$, $r$ & Number of experts, number of modules, LoRA rank (16) \\
        $d$, $n$ & Module input dimension, number of calibration tokens \\
        $\Delta W_t$, $U_t,\sigma_t,V_t$ & Expert update and its compact SVD (\cref{eq:svd}) \\
        $x_t$, $X_{t\ell}$ & Calibration image and its module input tokens \\
        $\hat\Sigma_{t\ell}$, $e_{t\ell,j}$ & Token second moment and its energy along read direction $v_{t\ell,j}$ \\
        $m_t$, $m_t^0$ & Margins (top-two gaps) on $x_t$ of the expert and pretrained model \\
        $\rho_t$, $\rho$ & Relative margin $m_t/m_t^0$ and its geometric mean over tasks \\
        $p$, $p_{\max}$ & Bank-wide spectral exponent (\cref{eq:p-rule}) and its ceiling $\frac34$ \\
        $p_t$ & Expert-specific exponent in the variant of \cref{sec:supp-mixed} \\
        $F_t(p)$ & Flattened target of the same Frobenius norm as $\Delta W_t$ \\
        $P_t(p)$, $\varepsilon$ & Weight input prior and its ridge $0.01$ \\
        $C_t(p)$, $g$ & Reconstruction metric and its image weight $\frac12$ \\
        $\kappa_{t\ell}$, $\kappa_t$ & Local response correction and its geometric mean over modules \\
        $\lambda_t$, $\nu$ & Target scale $\lambda_t=\nu\kappa_t$ and common factor $\nu=\frac12$ \\
        $E(A)$, $\|A\|_{\rm op}$ & Root mean square response of $A$ on the calibration tokens, and its peak gain \\
        $B(p)$ & Scaled target $\lambda_tF_{t\ell}(p)$ \\
        $\Delta W^\star$ & Merged update added to the pretrained weights \\
        \bottomrule
    \end{tabular}
    \caption{\textbf{Notation.} The index $t$ denotes an expert, $\ell$ a module and $j$ a singular direction.}
    \label{tab:supp-notation}
\end{table}

We assume that each expert update has rank exactly $r=16$, so its $r$ singular values are strictly positive.

\subsection{Closest flat spectrum}
\label{sec:supp-math-flat}
Full flattening should remove the dominance of the top direction while changing the expert as little as possible.
The target $F_t(0)$ does both.

\smallskip\noindent\textbf{Lemma 1.} \emph{Among the matrices whose $r$ nonzero singular values are equal and whose
Frobenius norm is $\|\Delta W_t\|_F$, the closest to $\Delta W_t$ is}
\begin{equation*}
    F_t(0)=a\,U_tV_t^\top,\qquad a=\frac{\|\sigma_t\|_2}{\sqrt r} .
\end{equation*}
This is the rank-$r$ form of a classical property of the polar factor $U_tV_t^\top$~\cite{fanhoffman,higham}. We
include the short proof.

\smallskip\noindent\emph{Proof.} Such a matrix $Z$ has $\|Z\|_{\rm op}=a$ and $\|Z\|_F=\|\Delta W_t\|_F$. Hence
\begin{equation*}
    \|Z-\Delta W_t\|_F^2=2\|\Delta W_t\|_F^2-2\langle Z,\Delta W_t\rangle_F ,
\end{equation*}
and, by \cref{eq:svd},
\begin{equation*}
    \langle Z,\Delta W_t\rangle_F=\sum_{j=1}^{r}\sigma_j\,u_j^\top Zv_j\;\le\;a\sum_{j=1}^{r}\sigma_j .
\end{equation*}
$F_t(0)$ attains this bound. Conversely, since every $\sigma_j>0$, equality forces $u_j^\top Zv_j=a=\|Z\|_{\rm op}$,
hence $Zv_j=au_j$, for every $j$. Then $\|ZV_t\|_F^2=ra^2=\|Z\|_F^2$, so $Z$ vanishes outside the span of $V_t$ and
$Z=aU_tV_t^\top$.\hfill$\square$

\smallskip Without the norm constraint, the same argument with the best scale gives $\bar\sigma_tU_tV_t^\top$, where $\bar\sigma_t$ is the
mean singular value. Iso-C applies this rule to the summed update~\cite{isoc}. \method instead flattens each expert
before merging, keeps its norm in $F_t$, and leaves its scale to $\lambda_t$ (\cref{sec:rescale}).

\subsection{Unique reconstruction}
\label{sec:supp-math-reconstruction}
One image spans at most $n$ input directions, so the merge must stay well defined outside them. The weight input
prior makes it unique and constrains it on the directions that the image misses.

The prior is the normalized Gram matrix of the target. By \cref{eq:target},
$F_t^\top F_t=c_t^2V_t\operatorname{diag}(\sigma_t^{2p})V_t^\top$ and $\|F_t\|_F^2=c_t^2\sum_j\sigma_{t,j}^{2p}$, so
\cref{eq:prior} reads
\begin{equation*}
    P_t=\frac{r\,F_t^\top F_t}{\|F_t\|_F^2}+\varepsilon I,\qquad \operatorname{tr}P_t=r+\varepsilon d .
\end{equation*}

\smallskip\noindent\textbf{Lemma 2.} \emph{For any scales $\lambda_t>0$, \cref{eq:objective} has a unique minimizer
$\Delta W^\star$, however few input directions the images span. Its rank is at most $\min(d_{\rm out},d,Tr)$.}

\smallskip\noindent\emph{Proof.} By \cref{eq:metric},
\begin{equation*}
    C_t=\frac{d\,\hat\Sigma_t}{2\operatorname{tr}\hat\Sigma_t}+\frac{d\,P_t}{2(r+\varepsilon d)} .
\end{equation*}
Since $\hat\Sigma_t\succeq0$, $\operatorname{tr}\hat\Sigma_t>0$, and $P_t\succeq\varepsilon I$, $C_t$ is symmetric and
\begin{equation*}
    C_t\;\succeq\;\frac{d\,P_t}{2(r+\varepsilon d)}\;\succeq\;\frac{\varepsilon d}{2(r+\varepsilon d)}\,I\;\succ\;0 ,
\end{equation*}
so $H=\sum_tC_t$ is positive definite. The objective
\begin{equation*}
    \mathcal R(Z)=\sum_{t=1}^{T}\operatorname{tr}\big((Z-\lambda_tF_t)\,C_t\,(Z-\lambda_tF_t)^\top\big)
\end{equation*}
is then strictly convex. Its gradient $2\big(ZH-\sum_t\lambda_tF_tC_t\big)$ vanishes only at $\Delta W^\star$, which
is thus the unique minimizer. Finally, each term $\lambda_tF_tC_tH^{-1}$ of $\Delta W^\star$ has its columns in the span of $U_t$, so the columns of
$\Delta W^\star$ lie in a space of dimension at most $Tr$.\hfill$\square$

\smallskip\noindent\emph{Missed directions.} Let $\xi$ be a unit input direction that the tokens miss,
$\tilde X_t^\top\xi=0$. With $G=\tilde X_t\tilde X_t^\top$ and $\tilde G_t=\alpha G+(1-\alpha)\operatorname{diag}(G)$,
RegMean (\cref{eq:regmean}) gives this direction the weight
\begin{equation*}
    \xi^\top\tilde G_t\,\xi=(1-\alpha)\sum_k G_{kk}\,\xi_k^2 .
\end{equation*}
It comes from the energies of the tokens along the coordinate axes, not from the read directions of the expert. In
\method, the prior alone gives $\xi$ the weight
\begin{equation*}
    \xi^\top C_t\,\xi\;\ge\;\frac{d}{2(r+\varepsilon d)}\,\xi^\top P_t\,\xi ,
\end{equation*}
which is largest along the read directions of the target.

\subsection{Response on the calibration image}
\label{sec:supp-math-scaling}
Flattening keeps the norm of a target but moves weight from its top read direction, which the image excites
strongly, to the weak ones (\cref{sec:metric}). We show that its response on the image then drops, so that
$\kappa_{t\ell}\ge1$, and drops more as the target gets flatter.

For a matrix $A$ acting on the module, let $E(A)=\|AX\|_F/\sqrt n$ be its root mean square response on the $n$
calibration tokens $X$, so that $E(A)^2=\operatorname{tr}(A\hat\Sigma A^\top)$. For $A=U\operatorname{diag}(s)V^\top$,
with the read directions $V$ of the expert and orthonormal columns in $U$,
\begin{equation*}
    E(A)^2=\sum_j s_j^2\,e_j,\qquad e_j=v_j^\top\hat\Sigma\,v_j .
\end{equation*}
Taking $s=\sigma$ and $s=c(p)\,\sigma^p$, \cref{eq:kappa} becomes
\begin{equation*}
    \kappa_{t\ell}(p)^2=\frac{f(1)}{f(p)},\quad f(p)=\sum_j q_p(j)\,e_j,\quad q_p(j)=\frac{\sigma_j^{2p}}{\sum_i\sigma_i^{2p}} .
\end{equation*}
Thus $f(p)$ is the mean energy of the tokens over the read directions, weighted by $\sigma_j^{2p}$. A flatter target
gives more weight to the weak directions. We assume $\sum_je_j>0$. Otherwise both responses vanish and
$\kappa_{t\ell}$ is undefined.

\smallskip\noindent\textbf{Proposition 1.} \emph{If $e_j\ge e_k$ whenever $\sigma_j>\sigma_k$, then
$\kappa_{t\ell}(p)\ge1$ on $[0,1]$ and $\kappa_{t\ell}$ is nonincreasing in $p$. The reverse ordering reverses both
conclusions.}

\smallskip\noindent\emph{Proof.} Differentiating the weights $q_p(j)$ gives
\begin{equation*}
    \begin{aligned}
        f'(p)&=2\operatorname{Cov}_{q_p}\!\big(e_j,\log\sigma_j\big)\\
             &=\sum_{j,k}q_p(j)\,q_p(k)\,(e_j-e_k)\log\frac{\sigma_j}{\sigma_k} ,
    \end{aligned}
\end{equation*}
whose terms are all nonnegative under the ordering. Thus $f$ is nondecreasing, $f(1)\ge f(p)$, and
$\kappa_{t\ell}^2=f(1)/f(p)$ is nonincreasing in $p$.\hfill$\square$

\smallskip The ordering is sufficient, not necessary, since the proof only uses $f'\ge0$. In line with Proposition 1, the median $\kappa_t$ is above 1 on every
bank, and largest on the fully flattened KnOTS banks, 1.75 on ViT-B/32 and 1.59 on ViT-L/14 (\cref{tab:supp-banks}).

\subsection{Why the margin sets the flattening}
\label{sec:supp-math-margin}
In \cref{sec:banks}, the top directions of the other experts cost 14 to 19 points on small-margin banks and nothing
on large-margin ones, and the best $p$ grows with the margin. Two simple results explain this. The margin bounds the
interference that a prediction tolerates (Lemma 3), and flattening lowers the largest interference that a target can
cause (Proposition 2).

\smallskip\noindent\textbf{Lemma 3.} \emph{Let $m$ be the gap between the two highest scores. If every score changes by
less than $m/2$, the predicted class stays the same. Changes of $m/2$ can create a tie.}

\smallskip\noindent\emph{Proof.} Changing each score by at most $u$ reduces any lead by at most $2u$. Lowering the top
score and raising the second by $m/2$ creates a tie.\hfill$\square$

\smallskip The margin $m_t$ is thus the tolerance of expert $t$ on its image, and $\rho_t$ compares it with the
tolerance of the pretrained model.

A target interferes with another task through the change it causes on its inputs. The largest change that a matrix
$B$ causes on a unit input is $\|B\|_{\rm op}$, which we call its peak gain. We compare it with the response $E(B)$ of
the target on its own image, which flattening should keep (\cref{sec:supp-math-scaling}).

\smallskip\noindent\textbf{Proposition 2.} \emph{Fix an expert and a module with $\sum_j e_j>0$. For $0\le p\le1$, let
$B(p)=\lambda_t(p)F_{t\ell}(p)$ be its scaled target. Then}
\begin{equation*}
    \frac{\|B(p)\|_{\rm op}^2}{E(B(p))^2}=\Big(\sum_j e_j\Big(\frac{\sigma_j}{\sigma_1}\Big)^{2p}\Big)^{-1},
\end{equation*}
\emph{which is nondecreasing in $p$. At $p=0$, it equals $(\sum_j e_j)^{-1}$, the smallest value over all matrices $A$
that read only the directions $V$, $A=AVV^\top$, with $E(A)>0$.}

\smallskip\noindent\emph{Proof.} Since $B(p)=\lambda_tc(p)\,U\operatorname{diag}(\sigma^p)V^\top$,
\begin{equation*}
    \begin{gathered}
        \|B(p)\|_{\rm op}=\lambda_tc(p)\,\sigma_1^p,\\
        E(B(p))^2=\lambda_t^2c(p)^2\sum_j\sigma_j^{2p}e_j ,
    \end{gathered}
\end{equation*}
whose ratio gives the identity. Each $(\sigma_j/\sigma_1)^{2p}\le1$ is nonincreasing in $p$. For $A=AVV^\top$,
\begin{equation*}
    E(A)^2=\operatorname{tr}\big(AV\,(V^\top\hat\Sigma V)\,V^\top A^\top\big)\;\le\;\|A\|_{\rm op}^2\sum_je_j .
\end{equation*}
Hence $\|A\|_{\rm op}^2/E(A)^2\ge(\sum_je_j)^{-1}$, with equality for $B(0)$.\hfill$\square$

\smallskip For the same response on its own image, the fully flattened target thus has the smallest peak gain.
Across modules, $E(B_\ell)/E(\Delta W_{t\ell})=\nu\kappa_t/\kappa_{t\ell}$ has geometric mean $\nu$
for every $p$, so the geometric mean of the peak gains $\|B_\ell(p)\|_{\rm op}$ is also smallest at $p=0$.

\smallskip\noindent\emph{Reading of \cref{sec:banks}.} Together, Lemma 3 and Proposition 2 explain the trade-off behind
\cref{eq:p-rule}. The target of another expert changes the module output on a unit input of task $t$ by at most its
peak gain. The prediction of task $t$ stays the same while the resulting score changes stay below $m_t/2$. In a bank
of small-margin experts, every prediction tolerates little interference, so every target should have the smallest
peak gain. These are the fully flattened targets, which \method chooses when $\rho\le1$. Experts with large margins
tolerate more, so their targets can keep part of their spectrum, whose top direction also carries their skill
(\cref{fig:direction1}). This matches \cref{fig:banks}b, where the best $p$ grows with the margin. The argument is
qualitative. It does not bound how a module output reaches the scores, and it does not give the form of
\cref{eq:p-rule}, which is empirical (\cref{sec:p-rule}).

\section{Experimental details}
\label{sec:supp-details}
\subsection{Banks}
\label{sec:supp-banks}
\begin{table}[H]
    \centering
    \footnotesize
    
    \setlength{\tabcolsep}{3pt}
    \begin{tabular}{@{}l p{6.2cm}@{}}
        \toprule
        LoRA~\cite{lora} & rank 16, $\alpha=16$, dropout 0.1, on the query, key, value and output projections of all
        blocks of the CLIP image encoder~\cite{clip} \\
        Head & frozen zero-shot CLIP head, logits $s$ times the cosine scores (\cref{eq:losses}) \\
        Optimizer & AdamW~\cite{adamw}, learning rate $3\cdot10^{-4}$, weight decay 0.1, batch size 32 \\
        Schedule & cosine~\cite{sgdr}, 500 warm-up steps, at most 50\,000 steps \\
        Early stopping & 10\% of the training split held out (at most 2\,000 images), evaluated every 2\,000 steps,
        best checkpoint, patience of five evaluations \\
        \bottomrule
    \end{tabular}
    \caption{\textbf{Training recipe of our banks.}}
    \label{tab:supp-recipe}
\end{table}

We train every bank except KnOTS with the recipe of \cref{tab:supp-recipe} and change only the loss. The KnOTS
banks are the public experts of~\cite{knots}.

CosFace~\cite{cosface} trains each expert so that the true class beats every other class by at least $\mu$. The margin
is used only in this loss, and the experts are merged and evaluated with the plain cosine head. The heads are frozen,
however, and no image feature can separate two nearly identical heads. For each class $y$, the gap between its score
and the best other score is therefore bounded, whatever the image, by a value $\gamma(y)$ that depends only on the
heads. The heads of the ten digits of MNIST~\cite{mnist} and SVHN~\cite{svhn} are so close that $\gamma(y)$ is at most
0.13 on ViT-B/32. A margin of 0.2 can thus never be reached for a digit. We therefore ask each class for at most
$0.8\,\gamma(y)$ in $\mathrm{CF}_{0.116}$, $\mathrm{CF}_{0.188}$ and $\mathrm{CF}_{0.231}$. We compute $\gamma(y)$ as
the distance from the head of $y$ to the convex hull of the other heads.

\begin{table}[H]
  \centering
  \footnotesize
  \setlength{\tabcolsep}{3.2pt}
  \begin{tabular}{@{}l c r c ccc c@{}}
    \toprule
    Bank & Loss & $s$ & $\mu$ & $\rho$ & $p$ & $\kappa_t$ & Experts \\
    \midrule
    \multicolumn{8}{@{}l}{\textit{ViT-B/32}} \\
    KnOTS & public & -- & -- & 0.35 & 0.00 & 1.75 & 84.0 \\
    $\mathrm{CE}_{0.004}$ & CE & 10\,000 & -- & 0.78 & 0.05 & 1.13 & 88.1 \\
    $\mathrm{CE}_{0.008}$ & CE & 3\,000 & -- & 1.51 & 0.22 & 1.06 & 88.5 \\
    $\mathrm{CE}_{0.014}$ & CE & 1\,000 & -- & 2.75 & 0.46 & 1.03 & 88.4 \\
    $\mathrm{CE}_{0.043}$ & CE & 300 & -- & 7.04 & 0.64 & 1.01 & 89.5 \\
    $\mathrm{CE}_{0.084}$ & CE & 100 & -- & 13.91 & 0.69 & 1.02 & 89.8 \\
    $\mathrm{CF}_{0.084}$ & CF & 10\,000 & 0.05 & 15.32 & 0.70 & 1.01 & 89.5 \\
    $\mathrm{CF}_{0.116}$ & CF & 10\,000 & 0.10$^\dagger$ & 19.44 & 0.71 & 1.01 & 89.6 \\
    $\mathrm{CF}_{0.188}$ & CF & 10\,000 & 0.20$^\dagger$ & 26.30 & 0.72 & 1.01 & 89.0 \\
    \midrule
    \multicolumn{8}{@{}l}{\textit{ViT-L/14}} \\
    KnOTS & public & -- & -- & 0.27 & 0.00 & 1.59 & 92.4 \\
    $\mathrm{CE}_{0.007}$ & CE & 10\,000 & -- & 0.42 & 0.00 & 1.22 & 92.6 \\
    $\mathrm{CE}_{0.097}$ & CE & 100 & -- & 8.00 & 0.65 & 1.02 & 93.9 \\
    $\mathrm{CF}_{0.100}$ & CF & 10\,000 & 0.05 & 7.98 & 0.65 & 1.02 & 93.7 \\
    $\mathrm{CF}_{0.231}$ & CF & 10\,000 & 0.20$^\dagger$ & 17.16 & 0.70 & 1.01 & 93.4 \\
    \bottomrule
  \end{tabular}
  \caption{\textbf{Expert accuracy and relative margins across training recipes.} $s$ and $\mu$ denote the training logit scale and CosFace margin ($^\dagger$capped per class). Over five image draws we report the geometric mean of $\rho$ and the arithmetic mean of $p$ (\cref{eq:p-rule}). The last two columns give median $\kappa_t$ and mean expert accuracy (\%) on their own tasks.}
  \label{tab:supp-banks}
\end{table}

On each backbone, our banks reach about the same accuracy on their own tasks but differ widely in confidence, with
relative margins $\rho$ (\cref{eq:rho}) from below 1 to above 17 (\cref{tab:supp-banks}). The accuracy of the experts
thus says little about how they merge. $\mathrm{CE}_{0.004}$ is as accurate as our other ViT-B/32 banks, but its experts
are less confident than
the pretrained model ($\rho<1$), and it is the bank that \method merges worst (\cref{tab:supp-clip-complete}).

\subsection{Images and splits}
\label{sec:supp-splits}
We use the following sets of images for each task.
\setlength{\leftmargini}{1.1em}\setlength{\leftmarginii}{1.1em}
\begin{itemize}\setlength{\itemsep}{1pt}\setlength{\parskip}{0pt}
    \item \splitname{train}, the training split, on which the experts are trained.
    \item Two sets of examples used to calibrate the merge.
    \begin{itemize}\setlength{\itemsep}{0pt}
        \item \splitname{calib\_fewshot}, unlabeled images. \method and the activation-based merges use one image per
        task from it, drawn five times with the same draws for all methods, and more images for the budgets of
        \cref{sec:budget}.
        \item \splitname{calib\_hparams}, 128 labeled images per task, on which the merges marked \ding{55} choose their
        hyperparameters.
    \end{itemize}
    \item \splitname{val}, up to 512 images per task, used to develop the method and for the
    diagnostics of \cref{sec:reconstruction,sec:margins}.
    \item \splitname{test}, 1\,536 images per task (1\,024 for DTD), 11\,776 in total, for the eight-task CLIP results.
    The exponent sweeps in \cref{fig:margin-rule,sec:supp-mixed} use test accuracy for retrospective comparisons.
    The reported merges select their settings before test evaluation.
\end{itemize}
The language and medical models have their own splits (\cref{sec:supp-nli,sec:supp-gemma}).

\subsection{Implementation of \method}
\label{sec:supp-impl}
\method follows \cref{alg:masar}. It merges each LoRA module separately, 48 modules of size $768\times768$ on
ViT-B/32 and 96 of size $1024\times1024$ on ViT-L/14. The rest of the encoder and the task heads stay those of the
pretrained model. For each task, one forward pass of the pretrained model on the unlabeled image gives $\hat\Sigma_t$ in
every module, over all tokens, and the scores of the pretrained model. One forward pass of the expert gives its scores.
The SVDs and the Cholesky solve of \cref{eq:objective} are computed in double precision. If the pretrained model gives
$m^0_t=0$, $\rho_t$ is undefined and task $t$ is left out of the geometric mean. This never happens on CLIP and happens
once on NLI (\cref{sec:supp-nli}). A very small $m^0_t$ gives a large $\rho_t$, but $p$ never exceeds $p_{\max}$. The
four constants, $\varepsilon=0.01$, the weight $g=\frac12$ of the image in \cref{eq:metric}, $p_{\max}=\frac34$ and
$\nu=\frac12$, are the same for all banks and models. \Cref{sec:supp-sensitivity} varies the first three, and
\cref{fig:lambda}b varies $\nu$. We use PyTorch on NVIDIA RTX A6000 GPUs.

\subsection{Baselines}
\label{sec:supp-baselines}
\begin{table}[H]
    \centering
    \footnotesize
    
    \setlength{\tabcolsep}{4pt}
    \begin{tabular}{@{}l l l@{}}
        \toprule
        Method & Chosen on labels & Values \\
        \midrule
        \multicolumn{3}{@{}l}{\textit{Tuned on labels}} \\
        TA, TSV-M, Iso-C, ACTMat & $\lambda$ & $\mathcal{G}$ \\
        ACE-Merging & $\lambda$ & $\mathcal{G}$ \\
        CT-Merging & $\gamma$ & $\mathcal{G}$ \\
        DC-Merge & smoothing, $\lambda$ & \{avg., linear\}${}\times\mathcal{G}$ \\
        Core Space & $\lambda$ & $\mathcal{G}'$ \\
        TIES, KnOTS-TIES & density, $\lambda$ & \{0.1, 0.2, \dots, 1\}${}\times\mathcal{G}'$ \\
        NSC & initial value & $\mathcal{G}$ up to 0.5 \\
        RegMean family$^+$ & $\lambda$ & $\mathcal{G}$ \\
        ReForge$^+$ & mix, scales, $\lambda$ & 200 trials \\
        \midrule
        \multicolumn{3}{@{}l}{\textit{Label-free}} \\
        RegMean, RegMean++ & -- & $\lambda=1$ \\
        CrossMean & -- & $\lambda=1$ \\
        AdaMerging, TARA-B & -- & 500 steps \\
        \textbf{\method} & -- & -- \\
        \midrule
        \multicolumn{3}{@{}l}{$\mathcal{G}=\{0.05, 0.1, 0.15, 0.2, 0.3, 0.4, 0.5, 0.6, 0.8, 1, 1.2, 1.5, 2, 2.5\}$} \\
        \multicolumn{3}{@{}l}{$\mathcal{G}'=\{0.02, 0.03, 0.05, 0.07, 0.1, 0.15, 0.2, 0.3, 0.4, 0.5, 0.6, 0.8, 1,$} \\
        \multicolumn{3}{@{}l}{\hphantom{$\mathcal{G}'=\{$}$1.25, 1.5, 2, 2.5, 3\}$} \\
        \bottomrule
    \end{tabular}
    \caption{\textbf{What each merge chooses on labels.} The merges tuned on labels keep the first maximum
    of the mean accuracy on \splitname{calib\_hparams} (128 labeled images per task).}
    \label{tab:supp-baselines}
\end{table}

\paragraph{Weight-only merges.} As in KnOTS~\cite{knots}, we evaluate every point of the grid
(\cref{tab:supp-baselines}) on \splitname{calib\_hparams} and evaluate only the chosen point on the test set. TIES~\cite{ties} and KnOTS-TIES use the code of KnOTS.
DC-Merge~\cite{dc_merge}, Core Space~\cite{corespace} and ACE-Merging~\cite{acemerging} use their official code, Core
Space with TSV followed by isotropic scaling and ACE-Merging with its ViT settings. We reimplement the other weight-only
merges from their papers, TA~\cite{taskarithmetic}, TSV-M~\cite{tsv}, Iso-C~\cite{isoc}, ACTMat~\cite{actmat} and
CT-Merging~\cite{ctmerging}. ACTMat and ACE-Merging infer the input statistics from the weights and have no
scale in their papers. Both were designed for fully fine-tuned experts, whose updates read every input direction,
whereas each of our LoRA updates reads only 16. At $\lambda=1$ they fall to 4 to 39\% on the banks of
\cref{tab:vitb32,tab:vitl14}, so we also choose their $\lambda$ on \splitname{calib\_hparams}.

\paragraph{Activation-based merges.} They use the same unlabeled image per task as \method and no label. RegMean~\cite{regmean},
RegMean++~\cite{regmeanpp} and CrossMean~\cite{crossmean} are closed form with $\lambda=1$, off-diagonal Gram entries scaled by $\alpha=0.9$ and a relative
ridge of $10^{-4}$. RegMean++ measures the inputs of each group of modules in the partially merged model, and CrossMean
uses the points $\{0,\frac12,1\}$ of its paper. AdaMerging~\cite{adamerging} starts its layer-wise coefficients at 0.3 and uses
Adam~\cite{adam} at $10^{-3}$. TARA-B~\cite{tara} starts at 0.4 and uses AdamW at $10^{-3}$ with the smooth Tchebycheff aggregation
of its paper. On CLIP, both run 500 steps of entropy minimization on batches of 16 images per task. With one image, the
batch repeats it, which is our only change to their recipes. NSC~\cite{nsc} follows its paper (100 steps of AdamW at
$10^{-3}$ on CLIP, 500 steps at $3\cdot10^{-4}$ on NLI and Gemma), but its coefficients barely move from their initial
value. The value of its paper, 0.4, is too large for the low-margin KnOTS experts (17.0 and 38.0\% on ViT-B/32 and
ViT-L/14), so we choose the initial value on \splitname{calib\_hparams} (\cref{tab:supp-baselines}).
The variants marked $^+$ in \cref{tab:supp-clip-complete} read the same image but use labels. RegMean,
RegMean++ and CrossMean then choose $\lambda$ on \splitname{calib\_hparams}. ReForge~\cite{reforge} uses its released
code, which mixes activation and weight Gram matrices, and scores the 200 trials of its Bayesian search over this mix, its
scales and $\lambda$ on \splitname{calib\_hparams}.

\subsection{Merging cost}
\label{sec:supp-cost}
\begin{table}[H]
  \centering
  \footnotesize
  \setlength{\tabcolsep}{2pt}
  \begin{tabular}{@{}l l rrrr rrr@{}}
    \toprule
    & & \multicolumn{4}{c}{Time (s)} & \multicolumn{3}{c}{Memory (GB)} \\
    \cmidrule(lr){3-6}\cmidrule(l){7-9}
    Method & Search & B/32 & L/14 & Llama & Gemma & B/32 & L/14 & Llama \\
    \midrule
    \multicolumn{9}{@{}l}{\textit{Tuned on labels}} \\
    TA & 14 points & 9.1 & 156 & 79.6 & 2{,}696 & 5.0 & 11.7 & 15.0 \\
    CT-Merging & 14 points & 10.1 & 159 & 85.5 & 2{,}859 & 5.0 & 11.7 & 15.0 \\
    DC-Merge & 28 points & 64.8 & 572 & 171 & -- & 5.0 & 11.7 & 15.0 \\
    NSC & 8 runs & 371 & 782 & 6{,}391 & 13{,}916 & 5.0 & 11.7 & 16.4 \\
    \midrule
    \multicolumn{9}{@{}l}{\textit{Label-free}} \\
    AdaMerging & 500 steps & 189 & 3{,}525 & -- & -- & 7.8 & 8.0 & -- \\
    TARA-B & 500 steps & 103 & 2{,}553 & 503 & 9{,}345 & 4.4 & 8.2 & 15.3 \\
    RegMean & -- & 0.3 & 1.6 & 752 & 221 & 6.3 & 16.9 & 18.1 \\
    RegMean++ & -- & 0.9 & 4.5 & 753 & 1{,}730 & 5.1 & 12.2 & 14.7 \\
    \oursrow \textbf{\method} & -- & 2.9 & 17.7 & 166 & 1{,}150 & 3.9 & 8.3 & 18.1 \\
    \bottomrule
  \end{tabular}
  \caption{\textbf{Without search or gradient steps, \method merges in seconds on CLIP.} Time to the final model and peak GPU memory, one unlabeled input per task. Search: grid points evaluated on labels (\cref{tab:supp-baselines}) or gradient steps. On CLIP and Llama-3-8B, \method solves on the CPU in double precision. One RTX A6000, two for Gemma-4-31B. --: not run on this model.}
  \label{tab:supp-cost}
\end{table}

\begin{table*}[t]
  \centering
  \footnotesize
  \setlength{\tabcolsep}{4pt}
  \begin{tabular}{@{}l ccccccccc c@{}}
    \toprule
    & KnOTS & $\mathrm{CE}_{0.004}$ & $\mathrm{CE}_{0.008}$ & $\mathrm{CE}_{0.014}$ & $\mathrm{CE}_{0.043}$ & $\mathrm{CE}_{0.084}$ & $\mathrm{CF}_{0.084}$ & $\mathrm{CF}_{0.116}$ & $\mathrm{CF}_{0.188}$ & Avg. \\
    \midrule
    Relative margin $\rho$ & 0.35 & 0.78 & 1.51 & 2.75 & 7.04 & 13.91 & 15.32 & 19.44 & 26.30 & -- \\
    Exponent $p$ of \method & 0.00 & 0.05 & 0.22 & 0.46 & 0.64 & 0.69 & 0.70 & 0.71 & 0.72 & -- \\
    \midrule
    \oursrow \textbf{\method} & 67.9 & 63.2 & 71.2 & 76.0 & 77.9 & 79.3 & 77.4 & 77.0 & 73.9 & 73.8 \\
    Prior flat, $p_P=0$ & 0.0 & 0.0 & 0.0 & $-0.3$ & $+0.1$ & $+0.1$ & $-0.3$ & $-0.1$ & $-0.1$ & $-0.1$ \\
    Target flat, $p_F=0$ & 0.0 & $+0.3$ & $+0.1$ & $-1.8$ & $-1.5$ & $-1.8$ & $-2.0$ & $-1.2$ & $-1.6$ & $-1.1$ \\
    Both flat & 0.0 & 0.0 & $-0.8$ & $-3.7$ & $-3.7$ & $-5.3$ & $-8.1$ & $-7.8$ & $-8.8$ & $-4.2$ \\
    \bottomrule
  \end{tabular}
  \caption{{\textbf{Adaptive flattening acts mostly through the target.} Test accuracy (\%) of \method on ViT-B/32, mean over five draws, and change in points when the exponent of the prior, of the target or of both is set to zero.}}
  \label{tab:supp-decoupling}
\end{table*}

On CLIP, most merges take much longer than \method because they search their hyperparameters, on a grid evaluated
on labels or with 500 gradient steps (\cref{tab:supp-cost}). \method and the RegMean family search nothing and merge in seconds.
\method is slower than RegMean mostly because our code solves in double precision on the CPU, which in turn keeps its GPU
memory below that of RegMean on CLIP.

\section{\texorpdfstring{Testing the label-free rules}{Testing the label-free rules}}
\label{sec:supp-rules}
\method sets its exponent and its scales from one unlabeled image, with constants fixed once for all banks. This section tests these rules. It first separates what the exponent changes (\cref{sec:supp-decoupling}), then varies the constants (\cref{sec:supp-sensitivity}) and the calibration image (\cref{sec:supp-calibration}). It then changes the setting rather than the method, with 2 to 16 experts (\cref{sec:supp-numexperts}) and with banks that mix two training recipes (\cref{sec:supp-mixed}), where one exponent per bank reaches its limit.

\subsection{Where margin-based flattening helps}
\label{sec:supp-decoupling}
\method uses one exponent for two roles, the flattening of the target $F_t$ and that of the weight prior $P_t$.
To see which role needs the adaptation to the margins, we give them separate exponents, $p_F$ for the target and $p_P$
for the prior, and set each one to zero (\cref{tab:supp-decoupling}).

Flattening both fully costs 4.2 points on average, and up to 8.8 points on the most confident bank. Flattening the
prior alone costs only 0.1 points, and the target alone 1.1. Adaptive flattening thus acts mostly through the target,
while the prior, even fully flattened, still completes the image with the directions that each expert reads.

\subsection{Sensitivity to the constants}
\label{sec:supp-sensitivity}
Besides the margin rule, \method has three constants fixed once for all banks, the ridge $\varepsilon$ of the
prior (\cref{eq:prior}), the weight $g$ of the image in the metric, and the ceiling $p_{\max}$ of the exponent
(\cref{eq:p-rule}). The scale factor $\nu$ is tested in \cref{fig:lambda}b. We vary each constant alone, the others
fixed, to check that no bank needs its own value.

\paragraph{Ridge $\varepsilon$.} We set the ridge $\varepsilon$ of \cref{eq:prior} to $10^{-2}$, so that the
prior gives some weight to every input direction, including those that neither the image nor the weights cover. We vary
$\varepsilon$ from $10^{-3}$ to $10^{-1}$ on five ViT-B/32 banks, the four of \cref{tab:vitb32} and $\mathrm{CE}_{0.008}$,
whose relative margin is close to 1, to see how much this protection matters (\cref{tab:supp-sensitivity}). It is
the constant that matters most. Ten times smaller, it costs 7.1 points on average and up to 11.2 on one bank, since the
merge then stops protecting these directions. A larger ridge helps the confident banks but hurts KnOTS, and
$\varepsilon=10^{-2}$ stays within 0.6 points of the best value tested on average.

\begin{table}[H]
  \centering
  \footnotesize
  \setlength{\tabcolsep}{1.6pt}
  \begin{tabular}{@{}l rrrrr r@{}}
    \toprule
    $\varepsilon$ & KnOTS & $\mathrm{CE}_{0.008}$ & $\mathrm{CE}_{0.014}$ & $\mathrm{CF}_{0.084}$ & $\mathrm{CF}_{0.188}$ & Avg. \\
    \midrule
    $10^{-3}$ & $-0.26$ & $-11.16$ & $-6.90$ & $-7.22$ & $-10.04$ & $-7.12$ \\
    $3\cdot10^{-3}$ & $+0.09$ & $-4.17$ & $-2.33$ & $-2.79$ & $-3.72$ & $-2.58$ \\
    \oursrow \textbf{\method}, $10^{-2}$ & 67.88 & 71.19 & 76.04 & 77.37 & 73.93 & 73.28 \\
    $3\cdot10^{-2}$ & $-0.94$ & $+1.21$ & $+0.24$ & $+0.93$ & $+1.51$ & $+0.59$ \\
    $10^{-1}$ & $-3.78$ & $-1.04$ & $-1.97$ & $+0.41$ & $+1.02$ & $-1.07$ \\
    \bottomrule
  \end{tabular}
  \caption{{\textbf{Sensitivity to the ridge $\varepsilon$.} Test accuracy (\%) of \method at $\varepsilon=10^{-2}$, mean over five draws of one unlabeled image per task. Other rows give the change in points with another ridge, computed with the same images.}}
  \label{tab:supp-sensitivity}
\end{table}

\paragraph{Image weight $g$.} More calibration images can cover more input directions, so we test whether the image moment should carry more weight. We vary $g$ in $C_t=g\,\mathcal{N}(\hat\Sigma_t)+(1-g)\,\mathcal{N}(P_t)$, with $g=\frac12$ as in \cref{eq:metric}. Surprisingly, giving the image moment more weight reduces accuracy even with more images per task (\cref{tab:supp-gbudget}). With one image per task, shifting some weight from the image moment to the prior yields a small gain.

\begin{table}[H]
  \centering
  \footnotesize
  \setlength{\tabcolsep}{2pt}
  \begin{tabular}{@{}l rrrrrr@{}}
    \toprule
    Images per task & 0 & 1 & 16 & 32 & 64 & 256 \\
    \midrule
    $g=0$ & 57.83 & $-11.66$ & $-15.34$ & $-15.63$ & $-15.78$ & $-15.85$ \\
    $g=0.10$ & -- & $-0.28$ & $-1.12$ & $-1.22$ & $-1.26$ & $-1.26$ \\
    $g=0.25$ & -- & $+0.24$ & $-0.10$ & $-0.12$ & $-0.10$ & $-0.10$ \\
    $g=0.40$ & -- & $+0.17$ & $+0.07$ & $+0.03$ & $+0.05$ & $+0.05$ \\
    \oursrow \textbf{\method}, $g=0.50$ & -- & 74.33 & 77.36 & 77.64 & 77.76 & 77.84 \\
    $g=0.75$ & -- & $-0.61$ & $-0.41$ & $-0.47$ & $-0.49$ & $-0.47$ \\
    $g=0.975$ & -- & $-3.13$ & $-3.13$ & $-3.38$ & $-3.61$ & $-3.75$ \\
    $g=1$ & -- & $-27.97$ & $-8.03$ & $-7.43$ & $-7.20$ & $-7.09$ \\
    \bottomrule
  \end{tabular}
  \caption{\textbf{Image weight $g$.} Test accuracy (\%) at $g=0.50$ and changes in points over four banks and ten draws. Images still set $p$ and $\lambda_t$ at $g=0$. The zero-image accuracy uses $p=0$ and $\lambda_t=1/2$. At $g=1$, we use the pseudoinverse.}
  \label{tab:supp-gbudget}
\end{table}

With 1 to 256 images per task, both $g=0$ and $g=1$ reduce accuracy. This supports combining the two terms in $C_t$, with $g=\frac12$ as a good compromise.

\paragraph{Exponent ceiling $p_{\max}$.} We chose $\frac34$ on validation to retain some flattening even on the most
confident banks. On the 14 banks, varying the ceiling from $\frac12$ to 1 gains at most 0.02 points on average
(\cref{tab:supp-pmax}). A ceiling of 1 helps on none of the banks. Lowering it to $\frac12$ helps ViT-L/14 on average
and hurts ViT-B/32.

\begin{table}[H]
  \centering
  \footnotesize
  \setlength{\tabcolsep}{5pt}
  \begin{tabular}{@{}l rrr@{}}
    \toprule
    Ceiling & ViT-B/32 (9) & ViT-L/14 (5) & All 14 \\
    \midrule
    $p_{\max}=\frac12$ & $-0.18$ & $+0.10$ & $-0.08$ \\
    $p_{\max}=\frac58$ & $-0.01$ & $+0.07$ & $+0.02$ \\
    \oursrow \textbf{\method}, $p_{\max}=\frac34$ & 73.76 & 84.30 & 77.53 \\
    $p_{\max}=\frac78$ & $-0.03$ & $-0.08$ & $-0.05$ \\
    $p_{\max}=1$ & $-0.14$ & $-0.19$ & $-0.16$ \\
    \bottomrule
  \end{tabular}
  \caption{{\textbf{Sensitivity to the exponent ceiling $p_{\max}$.} Test accuracy (\%) of \method at $p_{\max}=\frac34$, mean over five draws and over the banks of each backbone. Other rows give the change in points with another ceiling, computed with the same images.}}
  \label{tab:supp-pmax}
\end{table}

\subsection{The calibration image}
\label{sec:supp-calibration}

Since \method sees a single unlabeled image per task, we test whether it needs a typical image of the task, or
even an image of the task, by replacing the random image on the four ViT-B/32 banks of \cref{tab:vitb32} with images
chosen by expert confidence, images of other tasks of the bank, or external images (\cref{tab:supp-calibration-image}).

\begin{table}[H]
  \centering
  \footnotesize
  \setlength{\tabcolsep}{1.9pt}
  \begin{tabular}{@{}l ccccc@{}}
    \toprule
    Calibration image & KnOTS & $\mathrm{CE}_{0.014}$ & $\mathrm{CF}_{0.084}$ & $\mathrm{CF}_{0.188}$ & Avg. \\
    \midrule
    \oursrow \textbf{MASAR} (random image) & 68.1\std{0.6} & 76.3\std{0.6} & 77.8\std{0.5} & 74.3\std{0.6} & 74.1 \\
    \midrule
    Lowest confidence & $+0.1$ & $-0.5$ & $+0.1$ & $+0.3$ & $0.0$ \\
    Highest confidence & $-0.7$ & $+0.1$ & $-0.2$ & $-0.2$ & $-0.3$ \\
    \addlinespace[2pt]
    Another task in the bank & $-7.5$ & $-5.9$ & $-4.7$ & $-3.1$ & $-5.3$ \\
    External dataset & $-5.6$ & $-7.8$ & $-6.4$ & $-6.0$ & $-6.5$ \\
    \bottomrule
  \end{tabular}
  \caption{\textbf{One random image of the task is enough to calibrate \method.} Test accuracy (\%) of \method
  with one random unlabeled image per task, mean $\pm$ std over 64 draws per bank. Other rows give the change in points
  with another calibration image. \emph{Lowest} and \emph{highest confidence}: the image, among eight of the task, on
  which the expert has the smallest or largest gap between its two best scores. External images come from
  CIFAR-100~\cite{cifar100}, STL-10~\cite{stl10}, Food-101~\cite{food101} and Flowers-102~\cite{flowers102}.}
  \label{tab:supp-calibration-image}
\end{table}

With random images from the correct tasks, test accuracy is stable
(standard deviation of 0.5--0.6 points). Selecting images by expert
confidence changes the four-bank mean by at most 0.3 points.
Images from other tasks in the bank or from external datasets
lower accuracy by 5.3 and 6.5 points on average, respectively.
Overall, one random image from the correct task is enough to calibrate the merge.

\subsection{Number of experts}
\label{sec:supp-numexperts}
\begin{table}[H]
  \centering
  \footnotesize
  \setlength{\tabcolsep}{3.2pt}
  \begin{tabular}{w{l}{5.5em} w{c}{3em} *{6}{w{c}{\dimexpr(\linewidth-8.5em-16\tabcolsep)/6\relax}}}
    \toprule
    & & \multicolumn{4}{c}{Our eight tasks} & \multicolumn{2}{c}{Added tasks} \\
    \cmidrule(lr){3-6}\cmidrule(l){7-8}
    Experts $T$ & \shortstack{Label-\\free} & 2 & 4 & 6 & 8 & 12 & 16 \\
    \midrule
    \multicolumn{8}{@{}l}{\textit{Small margins} (KnOTS$^\ast$)} \\
    \oursrow \textbf{\method} & \checkmark & 73.7 & 72.3 & 70.1 & 67.9 & 66.9 & 65.1 \\
    DC-Merge & \ding{55} & 61.5 & 69.2 & 65.9 & 65.0 & 67.8 & 66.1 \\
    RegMean++ & \checkmark & 69.2 & 66.1 & 58.7 & 55.4 & 57.3 & 53.8 \\
    \addlinespace
    \multicolumn{8}{@{}l}{\textit{Large margins} ($\mathrm{CF}_{0.084}$, extended with the same loss)} \\
    \oursrow \textbf{\method} & \checkmark & 84.1 & 84.2 & 81.2 & 77.4 & 76.6 & 72.4 \\
    DC-Merge & \ding{55} & 70.6 & 78.9 & 76.5 & 73.3 & 73.8 & 72.0 \\
    RegMean++ & \checkmark & 83.5 & 77.3 & 70.1 & 63.9 & 66.8 & 66.1 \\
    \bottomrule
  \end{tabular}
  \caption{{\textbf{\method stays the best label-free merge from 2 to 16 experts.} Test accuracy (\%) on ViT-B/32, mean over tasks and, for $T<8$, five subsets. \method uses five one-image draws ($\lambda_t=\kappa_t/2$), RegMean++ uses the first ($\lambda=1$), and DC-Merge tunes its scale and smoothing on 128 labeled images per task. $^\ast$Public LoRAs~\cite{dc_merge} for $T=12,16$.}}
  \label{tab:supp-numexperts}
\end{table}

To test whether the constants of \method, set with eight experts, hold across the number of experts, we merge
from 2 to 16 ViT-B/32 experts, on a bank with small margins and on one with large margins (\cref{tab:supp-numexperts}). With 12 experts, we add CIFAR-100~\cite{cifar100},
Flowers102~\cite{flowers102}, Oxford-IIIT Pet~\cite{pets} and STL-10~\cite{stl10}, and with 16, also
CIFAR-10~\cite{cifar100}, FER2013~\cite{fer2013}, Fashion-MNIST~\cite{fashionmnist} and Rendered SST-2~\cite{clip}.
With small margins, the 12 and 16 experts are public rank-16 LoRAs~\cite{dc_merge}, trained on these tasks. For large margins we retain the eight $\mathrm{CF}_{0.084}$ experts and train
the additional experts with the same CosFace margin ($\mu=0.05$). At $T=12,16$, both banks use the same
test images and calibration draws.

\method stays the best label-free merge at every $T$, 11.3 points above RegMean++ with 16 small-margin
experts. With large margins, it also stays ahead of DC-Merge, tuned on labels, at every $T$, although its lead narrows
from 2.8 points at $T=12$ to 0.4 at $T=16$. With small margins, DC-Merge leads by 0.9 and 1.0 points at $T=12$ and 16,
the regime where \method already loses on the full banks (\cref{sec:supp-allbanks}). Every draw of the extended
CosFace banks gives $\rho>7$, so the rule keeps partial flattening there.

\subsection{Banks mixing two training recipes}
\label{sec:supp-mixed}

\method sets one exponent per bank, so every expert is flattened the same way. A bank may however mix hesitant
and confident experts, which call for opposite flattenings (\cref{sec:banks}). To test whether one exponent can serve
both, we build two banks on ViT-B/32 with four experts from KnOTS and four from $\mathrm{CF}_{0.188}$. We compare \method,
whose exponent comes from the mean relative margin of the bank, with a variant where each expert keeps its own adaptive
flattening, set by \cref{eq:p-rule} from its own relative margin $\rho_t$.

\begin{table}[H]
    \centering
    \footnotesize
    \setlength{\tabcolsep}{2.4pt}
    \begin{tabular}{@{}l ccc ccc@{}}
        \toprule
        & \multicolumn{3}{c}{Mix 1} & \multicolumn{3}{c}{Mix 2} \\
        \cmidrule(lr){2-4}\cmidrule(l){5-7}
        Method & All & KnOTS & CF & All & KnOTS & CF \\
        \midrule
        \oursrow \textbf{\method}, one $p$ per bank & 69.9 & 58.3 & 81.6 & 65.6 & 54.3 & 76.8 \\
        \method, one $p$ per expert & \textbf{72.9} & 64.4 & 81.5 & \textbf{66.5} & 56.5 & 76.6 \\
        \midrule
        DC-Merge~\cite{dc_merge}, labels & 70.4 & 62.0 & 78.8 & 63.9 & 51.9 & 76.0 \\
        \bottomrule
    \end{tabular}
    \caption{\textbf{When a bank mixes hesitant and confident experts, each expert needs its own exponent.}
    Test accuracy (\%) on ViT-B/32, mean over five draws, on all tasks and on the tasks of each group. Mix 1 takes the
    KnOTS experts of Cars, EuroSAT, RESISC45 and SUN397, Mix 2 those of the other four tasks.}
    \label{tab:supp-mixed}
\end{table}

\begin{figure}[H]
    \centering
    \includegraphics[width=0.8\linewidth]{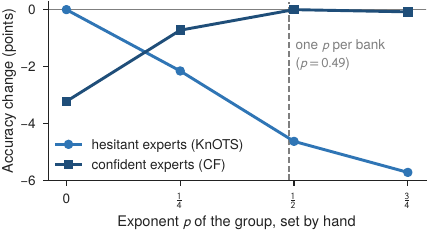}
    \caption{\textbf{Hesitant experts need $p=0$, confident experts $p\geq\frac12$.} Change in test accuracy
    on the tasks of each group as its exponent is set by hand, from its best exponent. Mean over both mixed banks.
    Dashed line, the single exponent of \method.}
    \label{fig:supp-mixed}
\end{figure}

One exponent per bank loses 3.0 and 1.0 points to one exponent per expert, all on the KnOTS tasks
(\cref{tab:supp-mixed}). To see why, we search on test labels the best exponent of each group among $0$, $\frac14$,
$\frac12$ and $\frac34$ (\cref{fig:supp-mixed}). The hesitant experts need $p=0$ and the confident ones $p\geq\frac12$, so
the single exponent of \method, about $\frac12$, suits only the confident experts and costs the hesitant ones 4.6 points.
An adaptive flattening per expert, $p_t$, lets each expert adjust its own spectrum, so that the hesitant experts are fully
flattened while the confident ones keep more of theirs.

On banks trained with one recipe, one exponent per expert loses 0.2 points on average
(\cref{tab:supp-own-main}), likely because the margin of one expert on one image is noisier than the mean over the bank.
One exponent per bank therefore remains the right choice when all experts share a recipe, as in the banks of the paper,
and an exponent per expert pays off only when a bank mixes recipes.

\begin{table}[H]
    \centering
    \footnotesize
    \setlength{\tabcolsep}{2.2pt}
    \begin{tabular}{@{}l rrrrr@{}}
        \toprule
        Method & KnOTS & $\mathrm{CE}_{0.014}$ & $\mathrm{CF}_{0.084}$ & $\mathrm{CF}_{0.188}$ & Avg. \\
        \midrule
        \oursrow \textbf{\method}, one $p$ per bank & 67.9 & 76.0 & 77.4 & 73.9 & 73.8 \\
        \method, one $p$ per expert & $+0.4$ & $-1.0$ & $-0.2$ & 0.0 & $-0.2$ \\
        \bottomrule
    \end{tabular}
    \caption{\textbf{On banks trained with one recipe, one exponent per expert does not help.} Test accuracy
    (\%) of \method, mean over five draws, and change in points with one exponent per expert, computed with the same
    images.}
    \label{tab:supp-own-main}
\end{table}

\section{Complete results on CLIP}
\label{sec:supp-clip}
\begin{table*}[p]
  \centering
  \footnotesize
  \renewcommand{\arraystretch}{0.94}
  \setlength{\tabcolsep}{1pt}
  \begin{tabular}{@{}>{\raggedright\arraybackslash}p{2.6cm} >{\centering\arraybackslash}p{0.95cm} >{\centering\arraybackslash}p{0.95cm} >{\centering\arraybackslash}p{\dimexpr(\textwidth-2.6cm-2\dimexpr0.95cm\relax-0.7cm-24\tabcolsep)/9\relax} >{\centering\arraybackslash}p{\dimexpr(\textwidth-2.6cm-2\dimexpr0.95cm\relax-0.7cm-24\tabcolsep)/9\relax} >{\centering\arraybackslash}p{\dimexpr(\textwidth-2.6cm-2\dimexpr0.95cm\relax-0.7cm-24\tabcolsep)/9\relax} >{\centering\arraybackslash}p{\dimexpr(\textwidth-2.6cm-2\dimexpr0.95cm\relax-0.7cm-24\tabcolsep)/9\relax} >{\centering\arraybackslash}p{\dimexpr(\textwidth-2.6cm-2\dimexpr0.95cm\relax-0.7cm-24\tabcolsep)/9\relax} >{\centering\arraybackslash}p{\dimexpr(\textwidth-2.6cm-2\dimexpr0.95cm\relax-0.7cm-24\tabcolsep)/9\relax} >{\centering\arraybackslash}p{\dimexpr(\textwidth-2.6cm-2\dimexpr0.95cm\relax-0.7cm-24\tabcolsep)/9\relax} >{\centering\arraybackslash}p{\dimexpr(\textwidth-2.6cm-2\dimexpr0.95cm\relax-0.7cm-24\tabcolsep)/9\relax} >{\centering\arraybackslash}p{\dimexpr(\textwidth-2.6cm-2\dimexpr0.95cm\relax-0.7cm-24\tabcolsep)/9\relax} >{\centering\arraybackslash}p{0.7cm}@{}}
    \toprule
    \textbf{ViT-B/32} & Images & Labels & KnOTS & $\mathrm{CE}_{0.004}$ & $\mathrm{CE}_{0.008}$ & $\mathrm{CE}_{0.014}$ & $\mathrm{CE}_{0.043}$ & $\mathrm{CE}_{0.084}$ & $\mathrm{CF}_{0.084}$ & $\mathrm{CF}_{0.116}$ & $\mathrm{CF}_{0.188}$ & Avg. \\
    \midrule
    Zero-shot & -- & -- & 48.2 & 48.2 & 48.2 & 48.2 & 48.2 & 48.2 & 48.2 & 48.2 & 48.2 & 48.2 \\
    Individual experts & -- & -- & 84.0 & 88.1 & 88.5 & 88.4 & 89.5 & 89.8 & 89.5 & 89.6 & 89.0 & 88.5 \\
    \midrule
    \multicolumn{13}{@{}l}{\textit{Weight-only ($\lambda$ chosen with labels)}} \\
    TA~\cite{taskarithmetic} & 0 & $\lambda$ & 51.4 & 52.3 & 59.4 & 69.3 & 75.8 & 75.8 & 76.1 & 74.4 & 69.2 & 67.1 \\
    TIES~\cite{ties} & 0 & $\lambda$ & 51.4 & 53.9 & 59.6 & 68.9 & 73.3 & 74.5 & 74.9 & 72.3 & 68.2 & 66.3 \\
    KnOTS-TIES~\cite{knots} & 0 & $\lambda$ & 56.1 & 56.5 & 62.8 & 71.3 & 75.2 & 74.7 & 74.6 & 73.4 & 69.2 & 68.2 \\
    TSV-M~\cite{tsv} & 0 & $\lambda$ & 54.0 & 59.0 & 65.7 & 72.7 & 76.3 & 77.4 & 76.4 & 76.0 & 72.1 & 70.0 \\
    Iso-C~\cite{isoc} & 0 & $\lambda$ & 62.0 & 62.6 & 68.2 & 70.4 & 70.4 & 73.9 & 70.8 & 73.4 & 69.7 & 69.1 \\
    DC-Merge~\cite{dc_merge} & 0 & $\lambda$ & 65.0 & \textbf{63.7} & 68.4 & 72.7 & 74.9 & 76.2 & 73.3 & 74.2 & 71.4 & 71.1 \\
    Core Space~\cite{corespace} & 0 & $\lambda$ & 62.9 & 63.0 & 68.8 & 69.8 & 69.9 & 73.6 & 70.7 & 72.6 & 69.1 & 68.9 \\
    CT-Merging~\cite{ctmerging} & 0 & $\gamma$ & 66.6 & 62.7 & 67.8 & 69.3 & 71.6 & 72.8 & 69.4 & 70.1 & 67.0 & 68.6 \\
    ACTMat~\cite{actmat} & 0 & $\lambda$ & 53.4 & 55.9 & 62.8 & 68.5 & 73.5 & 75.5 & 73.2 & 72.1 & 69.8 & 67.2 \\
    \midrule
    \multicolumn{13}{@{}l}{\textit{Activation-based (one unlabeled image per task)}} \\
    RegMean~\cite{regmean} & 1 & -- & 52.2\std{0.9} & 51.0\std{1.2} & 60.3\std{0.8} & 66.1\std{0.6} & 67.7\std{1.0} & 68.1\std{1.1} & 64.2\std{1.5} & 63.7\std{3.3} & 59.0\std{0.2} & 61.4 \\
    RegMean++~\cite{regmeanpp} & 1 & -- & 55.1\std{0.9} & 51.5\std{1.4} & 60.7\std{0.6} & 66.1\std{0.5} & 68.0\std{0.9} & 68.1\std{1.2} & 64.2\std{1.4} & 63.6\std{3.3} & 58.5\std{0.6} & 61.8 \\
    CrossMean~\cite{crossmean} & 1 & -- & 53.4\std{1.6} & 51.2\std{1.4} & 60.4\std{0.6} & 66.2\std{0.5} & 67.7\std{0.9} & 67.9\std{1.1} & 63.7\std{1.5} & 63.1\std{3.2} & 58.1\std{0.3} & 61.3 \\
    AdaMerging~\cite{adamerging} & 1 & -- & 57.6\std{0.7} & 57.5\std{1.1} & 60.1\std{1.6} & 66.1\std{1.7} & 74.8\std{0.6} & 73.9\std{1.7} & 73.8\std{1.4} & 72.6\std{1.7} & 67.6\std{0.9} & 67.1 \\
    TARA-B~\cite{tara} & 1 & -- & 57.3\std{1.2} & 58.5\std{2.5} & 66.9\std{2.0} & 72.5\std{1.0} & 75.5\std{0.6} & 74.8\std{0.4} & 75.0\std{0.2} & 72.0\std{1.1} & 60.9\std{2.0} & 68.2 \\
    NSC~\cite{nsc} & 1 & init. & 50.0\std{1.0} & 54.7\std{1.1} & 58.7\std{0.8} & 69.5\std{0.6} & 75.6\std{0.4} & 75.2\std{0.7} & 75.0\std{0.7} & 73.7\std{0.3} & 69.6\std{0.4} & 66.9 \\
    \oursrow \textbf{\method} & 1 & -- & \textbf{67.9}\std{0.7} & 63.2\std{0.5} & \textbf{71.2}\std{0.6} & \textbf{76.0}\std{0.8} & \textbf{77.9}\std{0.4} & \textbf{79.3}\std{0.2} & \textbf{77.4}\std{0.9} & \textbf{77.0}\std{0.6} & \textbf{73.9}\std{0.5} & \textbf{73.8} \\
    \midrule
    \multicolumn{13}{@{}l}{\textit{Additional merges tuned on labels, marked $^+$ (\cref{sec:supp-closest})}} \\
    \oursrow {\method}$^+$ & 1 & $\lambda$ & 67.8\std{0.8} & 66.7\std{0.8} & 72.0\std{0.5} & 75.8\std{0.6} & 77.6\std{0.4} & 79.2\std{0.3} & 77.6\std{0.7} & 77.2\std{0.4} & 75.4\std{0.3} & 74.4 \\
    RegMean~\cite{regmean}$^+$ & 1 & $\lambda$ & 56.4\std{0.4} & 56.5\std{1.2} & 61.6\std{0.5} & 65.8\std{0.7} & 67.5\std{1.0} & 68.1\std{0.8} & 65.2\std{1.2} & 64.7\std{2.9} & 62.9\std{1.0} & 63.2 \\
    RegMean++~\cite{regmeanpp}$^+$ & 1 & $\lambda$ & 58.4\std{0.7} & 57.6\std{0.9} & 61.5\std{0.7} & 66.0\std{0.6} & 67.7\std{1.1} & 67.9\std{0.9} & 64.9\std{1.3} & 65.1\std{2.3} & 62.4\std{1.2} & 63.5 \\
    CrossMean~\cite{crossmean}$^+$ & 1 & $\lambda$ & 56.6\std{0.8} & 55.7\std{1.2} & 61.1\std{0.5} & 66.2\std{0.6} & 67.8\std{1.3} & 68.0\std{1.2} & 63.7\std{1.5} & 63.1\std{3.3} & 58.7\std{1.3} & 62.3 \\
    ReForge~\cite{reforge}$^+$ & 1 & BO & 67.0\std{0.2} & 64.2\std{0.7} & 71.4\std{0.4} & 75.0\std{0.9} & 78.2\std{0.3} & 79.5\std{0.6} & 77.7\std{0.5} & 77.5\std{0.2} & 73.5\std{0.4} & 73.8 \\
    ACE-Merging~\cite{acemerging}$^+$ & 0 & $\lambda$ & 54.9 & 55.7 & 64.0 & 68.4 & 72.4 & 72.4 & 69.8 & 71.6 & 69.6 & 66.5 \\
    \bottomrule
  \end{tabular}
  \par\vspace{0.5em}
  \begin{tabular}{@{}>{\raggedright\arraybackslash}p{2.6cm} >{\centering\arraybackslash}p{0.95cm} >{\centering\arraybackslash}p{0.95cm} >{\centering\arraybackslash}p{\dimexpr(\textwidth-2.6cm-2\dimexpr0.95cm\relax-0.7cm-16\tabcolsep)/5\relax} >{\centering\arraybackslash}p{\dimexpr(\textwidth-2.6cm-2\dimexpr0.95cm\relax-0.7cm-16\tabcolsep)/5\relax} >{\centering\arraybackslash}p{\dimexpr(\textwidth-2.6cm-2\dimexpr0.95cm\relax-0.7cm-16\tabcolsep)/5\relax} >{\centering\arraybackslash}p{\dimexpr(\textwidth-2.6cm-2\dimexpr0.95cm\relax-0.7cm-16\tabcolsep)/5\relax} >{\centering\arraybackslash}p{\dimexpr(\textwidth-2.6cm-2\dimexpr0.95cm\relax-0.7cm-16\tabcolsep)/5\relax} >{\centering\arraybackslash}p{0.7cm}@{}}
    \toprule
    \textbf{ViT-L/14} & Images & Labels & KnOTS & $\mathrm{CE}_{0.007}$ & $\mathrm{CE}_{0.097}$ & $\mathrm{CF}_{0.100}$ & $\mathrm{CF}_{0.231}$ & Avg. \\
    \midrule
    Zero-shot & -- & -- & 65.2 & 65.2 & 65.2 & 65.2 & 65.2 & 65.2 \\
    Individual experts & -- & -- & 92.4 & 92.6 & 93.9 & 93.7 & 93.4 & 93.2 \\
    \midrule
    \multicolumn{9}{@{}l}{\textit{Weight-only ($\lambda$ chosen with labels)}} \\
    TA~\cite{taskarithmetic} & 0 & $\lambda$ & 68.5 & 67.7 & 87.2 & 86.4 & 84.7 & 78.9 \\
    KnOTS-TIES~\cite{knots} & 0 & $\lambda$ & 72.5 & 72.6 & 86.9 & 85.8 & 84.9 & 80.5 \\
    TSV-M~\cite{tsv} & 0 & $\lambda$ & 71.8 & 74.3 & 88.5 & 88.1 & 86.8 & 81.9 \\
    Iso-C~\cite{isoc} & 0 & $\lambda$ & 78.5 & 79.6 & 88.0 & 87.9 & 85.8 & 84.0 \\
    DC-Merge~\cite{dc_merge} & 0 & $\lambda$ & 78.4 & 79.0 & 88.5 & 86.5 & 87.0 & 83.9 \\
    Core Space~\cite{corespace} & 0 & $\lambda$ & 78.7 & 79.5 & 87.2 & 87.5 & 85.8 & 83.7 \\
    CT-Merging~\cite{ctmerging} & 0 & $\gamma$ & \textbf{80.2} & \textbf{79.7} & 87.1 & 87.0 & 86.4 & 84.1 \\
    ACTMat~\cite{actmat} & 0 & $\lambda$ & 70.7 & 73.1 & 88.2 & 87.9 & 85.9 & 81.2 \\
    \midrule
    \multicolumn{9}{@{}l}{\textit{Activation-based (one unlabeled image per task)}} \\
    RegMean~\cite{regmean} & 1 & -- & 70.2\std{1.8} & 65.2\std{0.7} & 82.9\std{0.8} & 82.1\std{0.9} & 79.7\std{1.0} & 76.0 \\
    RegMean++~\cite{regmeanpp} & 1 & -- & 71.1\std{1.6} & 66.4\std{0.6} & 83.1\std{0.7} & 82.3\std{0.9} & 79.7\std{1.0} & 76.5 \\
    CrossMean~\cite{crossmean} & 1 & -- & 70.2\std{1.6} & 63.6\std{1.8} & 80.5\std{0.9} & 79.5\std{1.1} & 75.4\std{1.0} & 73.8 \\
    TARA-B~\cite{tara} & 1 & -- & 78.5\std{1.1} & 79.0\std{1.3} & 87.2\std{0.2} & 87.0\std{0.3} & 83.5\std{0.2} & 83.1 \\
    NSC~\cite{nsc} & 1 & init. & 68.9\std{0.8} & 70.6\std{0.7} & 87.1\std{0.1} & 86.5\std{0.2} & 84.1\std{0.5} & 79.4 \\
    \oursrow \textbf{\method} & 1 & -- & 76.0\std{1.7} & 78.8\std{0.3} & \textbf{89.6}\std{0.1} & \textbf{88.9}\std{0.2} & \textbf{88.2}\std{0.1} & \textbf{84.3} \\
    \midrule
    \multicolumn{9}{@{}l}{\textit{Additional merges tuned on labels, marked $^+$ (\cref{sec:supp-closest})}} \\
    \oursrow {\method}$^+$ & 1 & $\lambda$ & 77.5\std{0.8} & 78.2\std{0.6} & 89.8\std{0.3} & 88.8\std{0.3} & 88.2\std{0.2} & 84.5 \\
    RegMean~\cite{regmean}$^+$ & 1 & $\lambda$ & 73.0\std{0.6} & 71.3\std{0.3} & 83.8\std{0.8} & 82.6\std{0.8} & 80.0\std{0.4} & 78.1 \\
    RegMean++~\cite{regmeanpp}$^+$ & 1 & $\lambda$ & 73.5\std{0.8} & 71.4\std{0.5} & 84.0\std{0.8} & 82.6\std{0.6} & 80.0\std{0.7} & 78.3 \\
    CrossMean~\cite{crossmean}$^+$ & 1 & $\lambda$ & 71.2\std{1.1} & 68.9\std{0.5} & 80.6\std{0.9} & 79.7\std{1.2} & 76.3\std{0.9} & 75.3 \\
    ReForge~\cite{reforge}$^+$ & 1 & BO & 80.2\std{0.4} & 82.7\std{0.4} & 89.7\std{0.3} & 89.2\std{0.5} & 87.3\std{0.4} & 85.8 \\
    ACE-Merging~\cite{acemerging}$^+$ & 0 & $\lambda$ & 72.0 & 76.0 & 88.2 & 87.8 & 85.8 & 82.0 \\
    \bottomrule
  \end{tabular}
  \caption{\textbf{On the 14 CLIP banks, \method is the best merge on 11 and the best label-free merge on 12.} Test accuracy (\%), mean $\pm$ std over five draws of the unlabeled image. \emph{Images}: unlabeled images per task. \emph{Labels}: what is chosen on 128 labeled images per task (init.: initial coefficient of NSC; BO: hyperparameters of ReForge); otherwise $\lambda=1$, and $\lambda_t=\frac12\kappa_t$ for \method. \emph{Avg.}: mean over the banks of the backbone. Bold: best merge, last block excluded.}
  \label{tab:supp-clip-complete}
  \label{tab:supp-allbanks}
  \label{tab:supp-closest}
\end{table*}

\begin{table*}[t]
  \centering
  \footnotesize
  \setlength{\tabcolsep}{3pt}
  \begin{tabular}{@{}l c cccccc c@{}}
    \toprule
    Method & \shortstack{Label-\\free} & SNLI & MNLI & SICK & QNLI & RTE & SciTail & Avg. \\
    \midrule
    Zero-shot &  & 41.9 & 34.7 & 53.1 & 51.9 & 42.0 & 60.6 & 47.4 \\
    Individual experts &  & 92.5 & 90.4 & 91.5 & 94.6 & 89.9 & 96.4 & 92.5 \\
    \midrule
    \multicolumn{9}{@{}l}{\textit{Weight-only ($\lambda$ chosen with labels)}} \\
    TA~\cite{taskarithmetic} & \ding{55} & 86.6 & 86.1 & 80.5 & 65.0 & 89.1 & 93.3 & 83.4 \\
    KnOTS-TIES~\cite{knots} & \ding{55} & 81.4 & 84.7 & \underline{84.4} & 78.2 & \textbf{90.6} & 94.0 & 85.6 \\
    TSV-M~\cite{tsv} & \ding{55} & 87.3 & \underline{87.3} & 74.0 & \underline{82.7} & 89.1 & 93.8 & 85.7 \\
    Iso-C~\cite{isoc} & \ding{55} & 84.1 & 82.8 & 78.2 & 72.9 & 89.9 & 94.2 & 83.7 \\
    DC-Merge~\cite{dc_merge} & \ding{55} & \textbf{89.5} & 86.7 & 81.6 & 73.5 & 89.1 & \underline{94.3} & \underline{85.8} \\
    Core Space~\cite{corespace} & \ding{55} & 83.6 & 84.9 & 76.7 & 76.2 & 89.9 & 92.6 & 84.0 \\
    CT-Merging~\cite{ctmerging} & \ding{55} & 82.7 & 67.7 & 75.1 & 82.0 & 78.3 & 88.8 & 79.1 \\
    ACTMat~\cite{actmat} & \ding{55} & 86.7 & 84.8 & 82.8 & 72.7 & \underline{90.6} & 94.1 & 85.3 \\
    \midrule
    \multicolumn{9}{@{}l}{\textit{Activation-based (one unlabeled example per task)}} \\
    RegMean~\cite{regmean} & \checkmark & \underline{89.5}\std{2.4} & 77.7\std{3.7} & 80.7\std{2.8} & 80.0\std{7.4} & 89.4\std{0.4} & 89.7\std{4.8} & 84.5 \\
    RegMean++~\cite{regmeanpp} & \checkmark & 88.9\std{2.4} & 78.5\std{3.2} & 81.6\std{2.8} & 82.2\std{6.2} & 89.4\std{0.4} & 90.2\std{4.5} & 85.1 \\
    CrossMean~\cite{crossmean} & \checkmark & 88.8\std{3.0} & 78.2\std{3.6} & 81.3\std{2.8} & 82.3\std{7.4} & 88.7\std{0.8} & 89.1\std{5.5} & 84.7 \\
    TARA-B$^\ddagger$~\cite{tara} & \checkmark & 80.3\std{6.4} & 82.4\std{5.0} & 50.7\std{5.9} & 57.4\std{6.9} & 88.0\std{0.6} & 91.5\std{0.6} & 75.0 \\
    NSC$^\ddagger$~\cite{nsc} & \checkmark & 86.9\std{0.5} & \textbf{87.7}\std{0.3} & 72.0\std{2.6} & 81.1\std{2.2} & 88.3\std{0.9} & 92.1\std{0.3} & 84.7 \\
    \midrule
    \oursrow \textbf{\method} (ours) & \checkmark & 88.2\std{0.4} & 84.2\std{1.3} & \textbf{85.1}\std{1.5} & \textbf{86.4}\std{1.9} & 90.3\std{0.6} & \textbf{95.9}\std{0.2} & \textbf{88.4} \\
    \midrule
    \multicolumn{9}{@{}l}{\textit{64 unlabeled examples per task}} \\
    RegMean~\cite{regmean} & \checkmark & 90.9\std{0.1} & 81.6\std{0.9} & 85.7\std{0.2} & 93.7\std{0.2} & 90.4\std{0.9} & 96.0\std{0.2} & 89.7 \\
    RegMean++~\cite{regmeanpp} & \checkmark & \textbf{91.7}\std{0.1} & 86.6\std{0.3} & \textbf{90.6}\std{0.3} & \textbf{94.5}\std{0.1} & \textbf{90.6}\std{1.1} & \textbf{96.4}\std{0.1} & \textbf{91.7} \\
    TARA-B$^\ddagger$~\cite{tara} & \checkmark & 75.1\std{3.7} & 77.4\std{2.5} & 46.3\std{1.6} & 50.5\std{0.2} & 87.7\std{1.3} & 91.3\std{0.4} & 71.4 \\
    NSC$^\ddagger$~\cite{nsc} & \checkmark & 86.7\std{0.1} & \textbf{87.5}\std{0.1} & 71.0\std{0.7} & 79.9\std{0.3} & 88.1\std{0.6} & 91.9\std{0.1} & 84.2 \\
    \oursrow \textbf{\method} (ours) & \checkmark & 90.8\std{0.0} & 80.0\std{0.3} & 88.2\std{0.2} & 90.7\std{0.2} & 86.4\std{0.6} & 94.1\std{0.4} & 88.4 \\
    \bottomrule
  \end{tabular}
  \caption{\textbf{\method transfers to Llama-3-8B unchanged.} Test accuracy (\%) of six NLI LoRAs merged into Llama-3-8B; mean $\pm$ std over five draws of the unlabeled example. The grid of $\lambda$ of Iso-C, DC-Merge and Core Space extends beyond 1.5 until the choice lies inside it. $^\ddagger$500 unsupervised steps (250 and three draws with 64 examples). Bold, underlined: best and second; last block apart.}
  \label{tab:supp-nli}
\end{table*}

\subsection{All 14 banks}
\label{sec:supp-allbanks}
On the 14 banks, \method is the best merge on 11 and the best label-free merge on 12
(\cref{tab:supp-clip-complete}). It loses only on small-margin banks, $\mathrm{CE}_{0.004}$, $\mathrm{CE}_{0.007}$ and
KnOTS ViT-L/14, by 0.5 to 4.2 points. The three banks where \method loses are those whose experts barely gain
confidence over the pretrained model ($\rho<1$, \cref{tab:supp-banks}). The exponent is not the cause, since
\cref{eq:p-rule} loses at most 0.2 points there against the best exponent of the grid (\cref{fig:margin-rule}). The
scale accounts for the loss on $\mathrm{CE}_{0.004}$ and for part of it on KnOTS ViT-L/14, since one $\lambda$ chosen on
labels at the same exponent gains 3.5 and 1.5 points there.

\subsection{Closest merges, tuned on labels}
\label{sec:supp-closest}
\method and the RegMean family read the same unlabeled image, but \method also scales each expert by $\lambda_t$, whereas
RegMean, RegMean++ and CrossMean have no scale in their papers. To show that the gain of \method does not come from this
scale, we rerun these three merges with a scale chosen on 128 labeled images per task, an advantage that \method does not
have. This scale is searched over $\mathcal{G}$ (\cref{tab:supp-baselines}).
We then observe in \cref{tab:supp-closest} that the scale adds at most 2.1 points on average, and that these
merges stay 6 to 12 points below \method on CLIP. With the same labels, \method gains only 0.2 to
0.6 points on average ({\method}$^+$), so our scales $\lambda_t$ are already close to the optimum. The gain of \method
thus comes from completing the image with the weights (\cref{sec:metric}). Only ReForge, which also completes the image
with the weights, matches or exceeds \method, but it tunes its hyperparameters on labels.

\section{\texorpdfstring{Transfer to other models}{Transfer to other models}}
\label{sec:supp-models}
\subsection{\texorpdfstring{\method on Llama-3, for NLI}{MASAR on Llama-3, for NLI}}
\label{sec:supp-nli}

\paragraph{Setup.} We check that the rules, set on CLIP image classifiers, carry over to text. We merge the six
NLI LoRA experts that KnOTS~\cite{knots} trained on Llama-3-8B~\cite{llama3}, for SNLI~\cite{snli}, MNLI~\cite{mnli},
SICK~\cite{sick}, QNLI and RTE~\cite{glue}, and SciTail~\cite{scitail} (rank 16 on the attention projections of the 32
blocks, one classification head each), and score each task with its own head on the KnOTS test split (29,216 rows).
One unlabeled training example per task, a premise and its hypothesis, replaces the image, and $\hat\Sigma_t$ is the
second moment of each module input over the tokens of this example. The heads read RMS-normalized states and $\rho_t$ cancels their common scale, so the
margins use the raw logits. Everything else is unchanged.

\paragraph{Results.} \method is the most consistent merge across tasks and draws (\cref{tab:supp-nli}). It is never
more than 3.5 points behind the best merge on any task, and it is four times more stable over draws than the
RegMean family. More examples help the merges that reconstruct the raw experts, and
with 64 examples per task, RegMean and RegMean++ overtake \method (89.7 and 91.7 against 88.4\%).

\paragraph{With a labeled scale.} As on CLIP (\cref{sec:supp-closest}), we give RegMean, RegMean++ and CrossMean a
scale chosen on 128 labeled examples per task. Unlike on CLIP, this closes their gap. They reach 87.7 to 88.4\%, level
with \method (88.4\%), and ReForge, tuned on the same labels, reaches 89.0\%.

\begin{table}[H]
  \centering
  \footnotesize
  \setlength{\tabcolsep}{5pt}
  \begin{tabular}{@{}l c c@{}}
    \toprule
    Calibration text (one per task) & Avg. & $p$ \\
    \midrule
    Premise and hypothesis of one example & 88.4\std{0.5} & 0.38--0.65 \\
    Premise and hypothesis of two examples & 88.4\std{0.6} & 0.30--0.49 \\
    Two random sentences & 85.5\std{2.3} & 0.00--0.47 \\
    \bottomrule
  \end{tabular}
  \caption{\textbf{Any premise and hypothesis of the task calibrate \method, random sentences do not.} Test accuracy (\%), mean $\pm$ std over five draws, and range of $p$. Random sentences: two GSM8K~\cite{gsm8k} questions.}
  \label{tab:supp-nli-ablation}
\end{table}

\paragraph{Which text calibrates.} We replace the calibration example of each task by a premise and a hypothesis
taken from two different examples of the task, or by two random sentences (\cref{tab:supp-nli-ablation}). The premise
and hypothesis of two examples calibrate as well as those of one example, so \method
only needs text from the task. Random sentences cost 2.8 points and make the merge four times less stable.

\renewcommand{\dbltopfraction}{0.99}
\setlength{\dblfloatsep}{8pt plus 2pt minus 2pt}\setlength{\dbltextfloatsep}{12pt plus 2pt minus 4pt}

\subsection{\texorpdfstring{\method on Gemma-4, for medical classification and VQA}{MASAR on Gemma-4, for medical classification and VQA}}
\label{sec:supp-gemma}

\paragraph{Setup.} We move further from CLIP, to Gemma-4-31B-it~\cite{gemma4}, a vision-language model
that answers in text, with seven medical LoRA experts of rank 16 on the attention and MLP projections of its 60
language blocks. Four classify images, for BreakHis~\cite{breakhis}, TBX11K~\cite{tbx11k}, LAG~\cite{lag} and
DDR~\cite{ddr}. Three do visual question answering (VQA)~\cite{vqa}, where the model answers a free-form question about
an image in a few words, for PathVQA~\cite{pathvqa}, SLAKE~\cite{slake} and VQA-Med 2019~\cite{vqamed}, and also adapt
the vision projector. All answer the same prompt in JSON. We report accuracy
for classification and normalized exact match for VQA, on 300 test rows per task stratified by class or answer type,
and 100 other rows per task, with other images, set up the adaptation below.

\begin{table*}[t]
  \centering
  \footnotesize
  \setlength{\tabcolsep}{3pt}
  \begin{tabular}{@{}l c ccccccc c@{}}
    \toprule
    Method & \shortstack{Label-\\free} & BreakHis & TBX11K & LAG & DDR & PathVQA & SLAKE & VQA-Med & Avg. \\
    \midrule
    Zero-shot &  & 56.3 & 53.7 & 76.0 & 56.3 & 37.3 & 53.3 & 29.7 & 51.8 \\
    Individual experts &  & 88.0 & 89.3 & 92.7 & 71.0 & 40.0 & 64.0 & 36.0 & 68.7 \\
    \midrule
    \multicolumn{10}{@{}l}{\textit{Weight-only ($\lambda$ chosen with labels)}} \\
    TA~\cite{taskarithmetic} & \ding{55} & 64.7 & 55.3 & 80.0 & 66.0 & 39.7 & 57.7 & 33.0 & 56.6 \\
    KnOTS-TIES~\cite{knots} & \ding{55} & 67.3 & 56.7 & 78.7 & 66.3 & 38.3 & 58.3 & 33.7 & 57.0 \\
    TSV-M~\cite{tsv} & \ding{55} & 75.0 & 64.0 & 84.7 & 65.7 & 36.0 & 59.7 & 33.7 & 59.8 \\
    Core Space~\cite{corespace} & \ding{55} & 69.7 & \textbf{93.3} & 87.0 & 67.7 & 35.3 & \textbf{62.0} & 33.7 & \textbf{64.1} \\
    CT-Merging~\cite{ctmerging} & \ding{55} & 73.3 & 72.0 & 88.3 & \underline{68.0} & 36.0 & 59.7 & 27.0 & 60.6 \\
    ACTMat~\cite{actmat} & \ding{55} & 60.0 & 54.0 & 83.0 & 61.3 & 40.0 & 57.7 & 31.3 & 55.3 \\
    \midrule
    \multicolumn{10}{@{}l}{\textit{Activation-based (one unlabeled image per task)}} \\
    RegMean~\cite{regmean} & \checkmark & 60.7\std{17.4} & 66.9\std{2.9} & 81.4\std{4.8} & 65.4\std{1.7} & \underline{40.4}\std{0.4} & \underline{60.4}\std{1.4} & \textbf{34.6}\std{0.8} & 58.6 \\
    RegMean++~\cite{regmeanpp} & \checkmark & 60.9\std{17.9} & 67.8\std{1.2} & 81.2\std{5.0} & 65.6\std{1.5} & \textbf{40.8}\std{0.7} & 60.2\std{1.0} & \textbf{34.6}\std{0.7} & 58.7 \\
    TARA-B$^\ddagger$~\cite{tara} & \checkmark & 71.3 & 59.7 & 85.0 & \textbf{69.0} & 39.7 & 57.7 & \underline{34.0} & 59.5 \\
    NSC$^\ddagger$~\cite{nsc} & \ding{55} & 65.4\std{0.8} & 55.0\std{0.6} & 78.9\std{1.5} & 65.8\std{0.2} & 39.8\std{0.4} & 56.9\std{0.5} & 33.2\std{0.2} & 56.4 \\
    \midrule
    \oursrow \textbf{\method} (ours) & \checkmark & \underline{75.4}\std{9.2} & \underline{85.7}\std{4.5} & \underline{90.3}\std{1.5} & 65.6\std{1.3} & 37.1\std{0.8} & 59.8\std{0.8} & 32.1\std{3.0} & \underline{63.7} \\
    \oursrow \method, first question only & \checkmark & \textbf{75.6}\std{9.0} & \underline{85.7}\std{4.7} & \textbf{90.4}\std{1.3} & 65.4\std{1.3} & 37.0\std{0.7} & 59.3\std{0.6} & 22.8\std{10.3} & 62.3 \\
    \midrule
    \multicolumn{10}{@{}l}{\textit{64 unlabeled images per task}} \\
    RegMean~\cite{regmean} & \checkmark & 79.0 & 56.0 & 84.0 & \textbf{68.0} & \textbf{40.3} & 63.7 & \textbf{35.3} & 60.9 \\
    RegMean++~\cite{regmeanpp} & \checkmark & \textbf{84.7} & 76.0 & 86.0 & 67.7 & 40.0 & \textbf{64.3} & 35.0 & \textbf{64.8} \\
    TARA-B$^\ddagger$~\cite{tara} & \checkmark & 72.7 & 56.0 & 76.7 & 67.3 & \textbf{40.3} & 59.7 & 33.3 & 58.0 \\
    \oursrow \textbf{\method} (ours) & \checkmark & 75.7 & \textbf{77.3} & \textbf{92.7} & 66.0 & 38.3 & 62.0 & 34.7 & 63.8 \\
    \bottomrule
  \end{tabular}
  \caption{\textbf{On seven medical experts, \method is the best label-free merge.} Accuracy (classification) and normalized exact match (VQA) of Gemma-4-31B-it with seven medical LoRA experts merged, on 300 test rows per task; mean $\pm$ std over three draws of the unlabeled image. Weight-only merges choose $\lambda$ on 64 labeled rows per task. \emph{First question only}: one question per calibration image instead of all. $^\ddagger$250 unsupervised steps, one draw; NSC, 500 steps and three draws, its initial coefficient chosen on the same labeled rows. Bold, underlined: best and second; last block: one draw, ranked apart.}
  \label{tab:supp-gemma-mix}
\end{table*}

\paragraph{Adaptation to generation.} Without a classification head, we score candidate answers by their
log-probability after the prompt and a format prefix taken from the expert's greedy answer. The candidates are the
classes for classification and, for VQA, the two best distinct answers of the expert and of the pretrained model,
found by beam search. The margin is the gap between the two best scores, averaged over the questions of the image, and
$\hat\Sigma_t$ is the second moment over the tokens of the prompt, the prefix and the answer. For VQA, the calibration
example is one unlabeled training image with all its questions. Using all of them helps, since with only the first question, VQA-Med falls from 32.1 to 22.8\% (\cref{tab:supp-gemma-mix}).

\paragraph{Results.} \method is the best label-free merge and is level with Core Space, the best merge tuned on
labels (\cref{tab:supp-gemma-mix}). Its lead comes from classification, whereas RegMean and
RegMean++ stay up to 3.7 points ahead on VQA. On a generative model, the rules thus still give the best label-free
merge, through the tasks that classify.

\subsection{\texorpdfstring{\method on LLaVA-1.5, for general multimodal tasks}{MASAR on LLaVA-1.5, for general multimodal tasks}}
\label{sec:supp-llava}
\begin{table*}[p]
  \centering
  \footnotesize
  \setlength{\tabcolsep}{2.5pt}
  \begin{tabular}{@{}l c *{9}{c}@{}}
    \toprule
    Method & \shortstack{Label-\\free} & SciQA & ImageNet & VQAv2 & REC & OCR & VizWiz & Flickr & IconQA & Avg. \\
    \midrule
    Zero-shot &  & 70.1 & 49.5 & 67.5 & 58.8 & 64.9 & 39.6 & 41.4 & 18.8 & 51.3 \\
    Individual experts &  & 89.0 & 95.5 & 67.9 & 70.9 & 65.5 & 64.9 & 57.3 & 75.4 & 73.3 \\
    \midrule
    \multicolumn{11}{@{}l}{\textit{Weight-only merges}} \\
    TA~\cite{taskarithmetic} & \ding{55} & 82.8 & 72.3 & 65.8 & 49.3 & 58.8 & 46.0 & 54.6 & 49.4 & 59.9 \\
    KnOTS-TIES~\cite{knots} & \ding{55} & 82.0 & 71.1 & 65.7 & 50.9 & 59.9 & 46.1 & 54.3 & 49.1 & 59.9 \\
    TSV-M~\cite{tsv} & \ding{55} & 81.3 & 69.2 & 66.9 & 53.0 & 60.9 & 46.6 & 53.7 & 47.4 & 59.9 \\
    Core Space~\cite{corespace} & \ding{55} & 80.7 & 64.2 & \textbf{67.7} & 52.6 & 62.5 & 46.3 & 51.5 & 44.2 & 58.7 \\
    CT-Merging~\cite{ctmerging} & \ding{55} & 81.0 & 65.7 & \underline{67.6} & 54.7 & 63.5 & 47.2 & 53.3 & 46.0 & 59.9 \\
    ACTMat~\cite{actmat} & \ding{55} & 83.7 & 72.6 & 65.6 & 50.3 & 56.0 & 44.9 & \textbf{55.3} & 51.3 & 60.0 \\
    DC-Merge~\cite{dc_merge} & \ding{55} & \textbf{84.8} & \textbf{75.0} & 65.4 & 46.2 & 56.3 & 44.6 & \underline{54.9} & \textbf{53.2} & 60.0 \\
    \midrule
    \multicolumn{11}{@{}l}{\textit{Activation-based (one unlabeled image per task)}} \\
    RegMean~\cite{regmean} & \checkmark & 77.8\std{0.7} & 64.9\std{0.4} & 67.6\std{0.2} & \textbf{62.2}\std{0.2} & \textbf{65.5}\std{0.1} & 47.1\std{0.3} & 53.6\std{0.2} & 39.9\std{1.8} & 59.8\std{0.3} \\
    RegMean++~\cite{regmeanpp} & \checkmark & 78.0\std{0.9} & 65.8\std{0.7} & 67.5\std{0.3} & \underline{61.9}\std{0.6} & \underline{65.3}\std{0.3} & 46.9\std{0.4} & 53.8\std{0.2} & 41.8\std{2.6} & 60.1\std{0.5} \\
    TARA-B~\cite{tara} & \checkmark & 82.6 & 72.0 & 65.9 & 49.3 & 60.9 & 45.9 & 53.9 & 51.3 & 60.2 \\
    NSC~\cite{nsc} & \ding{55} & \underline{84.2}\std{0.2} & \underline{72.8}\std{0.2} & 65.6\std{0.1} & 50.9\std{0.8} & 58.0\std{0.3} & 46.2\std{0.0} & 53.6\std{0.2} & \underline{51.6}\std{0.4} & 60.4\std{0.1} \\
    \midrule
    \oursrow \textbf{\method} (ours) & \checkmark & 82.6\std{0.2} & 67.8\std{1.2} & 67.2\std{0.2} & 58.0\std{0.7} & 64.5\std{0.2} & \textbf{47.4}\std{0.2} & 53.9\std{0.1} & 50.5\std{0.2} & \textbf{61.5}\std{0.1} \\
    \oursrow \method, first question only & \checkmark & 82.6\std{0.2} & 67.8\std{1.2} & 67.1\std{0.1} & 57.8\std{0.6} & 63.8\std{0.5} & \underline{47.4}\std{0.2} & 53.8\std{0.2} & 50.4\std{0.3} & \underline{61.3}\std{0.2} \\
    \midrule
    \multicolumn{11}{@{}l}{\textit{64 unlabeled images per task}} \\
    RegMean~\cite{regmean} & \checkmark & 78.9 & 64.9 & \textbf{67.9} & 61.9 & \textbf{65.6} & 47.3 & 53.8 & 42.2 & 60.3 \\
    RegMean++~\cite{regmeanpp} & \checkmark & 78.8 & \textbf{67.4} & 67.7 & \textbf{62.0} & 65.5 & 47.2 & \textbf{54.1} & 46.4 & 61.1 \\
    \oursrow \textbf{\method} (ours) & \checkmark & \textbf{82.6} & 67.3 & 67.2 & 58.4 & 64.4 & \textbf{47.5} & 53.7 & \textbf{50.5} & \textbf{61.4} \\
    \bottomrule
  \end{tabular}
  \caption{\textbf{On eight experts for general multimodal tasks, \method is the best merge.} Scores of LLaVA-v1.5-7B~\cite{llava15} with the eight LoRA experts of MM-MergeBench~\cite{robustmerge} merged, with the evaluation of RobustMerge; mean $\pm$ std over three draws of the unlabeled image, one for TARA-B and the 64-image block. Weight-only scales and the NSC initialization use 64 labeled rows per task. Bold: best merge of each budget; underlined: second with one image.}
  \label{tab:supp-llava}
\end{table*}

\begin{figure*}[p]
  \centering
  \includegraphics[width=\linewidth]{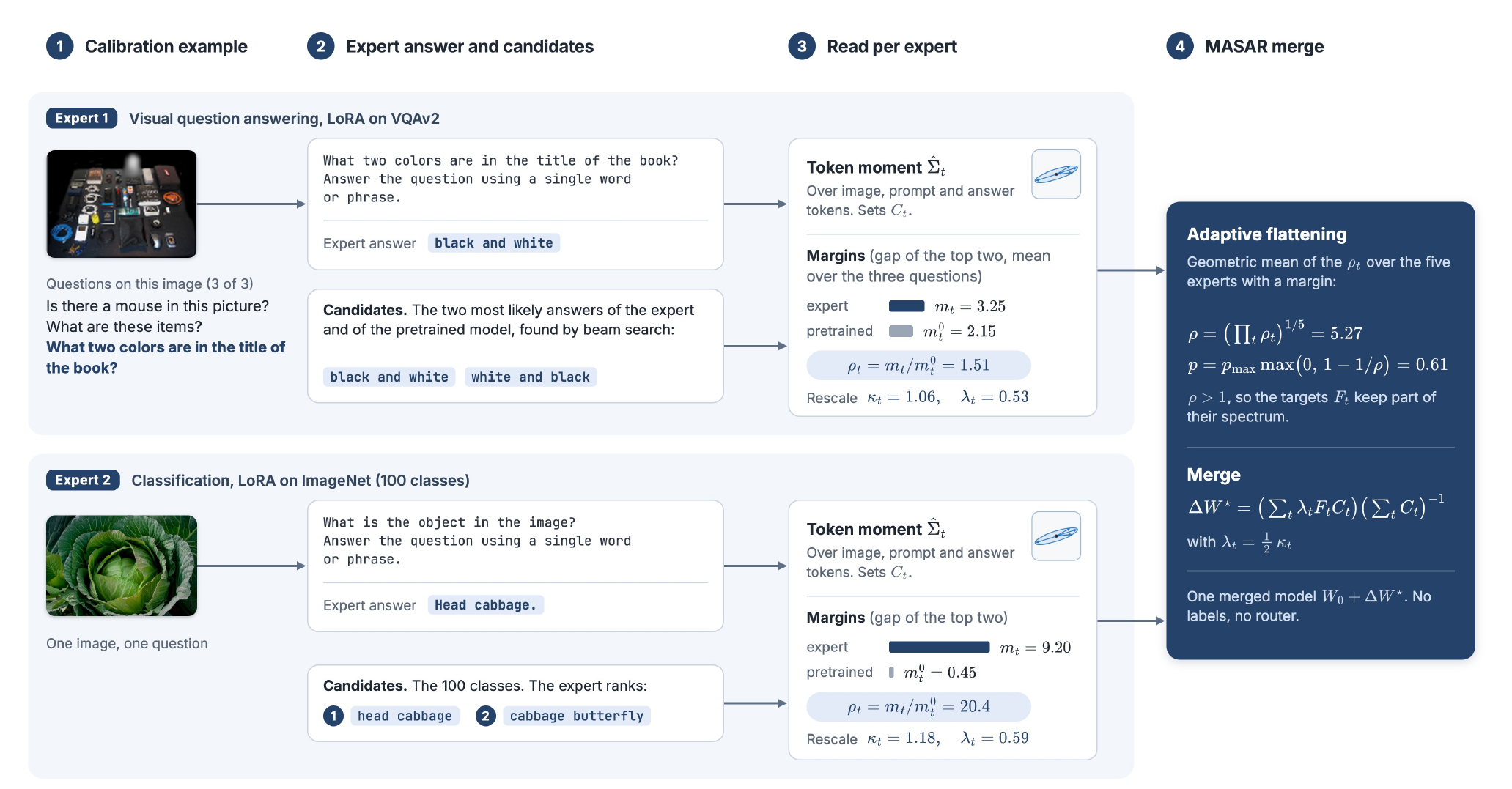}
  \caption{\textbf{What \method reads on the calibration example of each expert.} Two of the eight LLaVA-v1.5-7B experts
  of MM-MergeBench, first of the three draws of \cref{tab:supp-llava}. (1) Each expert gets one unlabeled image, with its
  questions for VQA. (2) It answers in the prompt of the benchmark. The candidates are the classes, or for VQA the best
  answers of a beam search with the expert and the pretrained model. (3) \method reads the token moment, which sets
  $C_t$, and the margin between the two best candidates, which gives $\rho_t$. (4) The ratios $\rho_t$ of the five
  experts with a margin set $p$, and \method merges. In the other two draws, $\rho<1$ and $p=0$.}
  \label{fig:supp-llava-calib}
\end{figure*}
\paragraph{Setup.} The experts of \cref{sec:supp-gemma} are medical, and most of their tasks have a
closed set of answers. We also merge the eight LoRA experts of rank 16 of MM-MergeBench~\cite{robustmerge}, a public
benchmark for merging multimodal experts, trained on LLaVA-v1.5-7B~\cite{llava15}, a smaller vision-language model than
Gemma-4-31B. Its tasks are general and cover several output formats. ScienceQA~\cite{scienceqa}
and IconQA~\cite{iconqa} are multiple-choice, ImageNet~\cite{imagenet} is classification, VQAv2~\cite{vqav2} and
OCR-VQA~\cite{ocrvqa} call for short open answers, REC for the bounding box of the object that a phrase describes on
RefCOCO~\cite{refcoco}, and VizWiz~\cite{vizwiz} and Flickr30k~\cite{flickr30k} for captions.
We score each task on its test file of MM-MergeBench, 59,088 examples in total, with the metrics of RobustMerge,
accuracy or exact match for most tasks, accuracy at IoU 0.5 for REC and the mean of the caption scores for VizWiz and
Flickr30k. The 264 ImageNet images that serve for calibration are left out of its test file. Each expert also tunes the
vision projector, and
every merge averages the projectors, as in RobustMerge. Margins are computed as in \cref{sec:supp-gemma} (\cref{fig:supp-llava-calib}). The
grounding expert answers with a box and the two captioning experts with a sentence, so they have no candidates and no
margin. They are merged like the other experts, and $\rho$ is the geometric mean over the five experts that have a
margin. The benchmark thus also tests \method when only some experts have a margin.

\paragraph{Results.} Although only five of the eight experts have a margin, \method is the best merge with one
unlabeled image per task, 1.3 points above the best label-free baseline, TARA-B, and 1.1 above the best merge that uses
labels, NSC (\cref{tab:supp-llava}). Its lead comes from the tasks with a closed set of answers, such as IconQA (50.5
against 41.8 for RegMean++), whereas RegMean++ keeps more of the grounding task (61.9 against 58.0). With 64 images per
task, \method stays ahead of RegMean++ (61.4 against 61.1).

\end{document}